\documentclass[sigconf]{acmart}
\usepackage{bm}
\usepackage{array}
\usepackage{multirow}
\usepackage{multicol}
\usepackage{graphicx}  
\usepackage{float}  
\usepackage{subfigure}  

\copyrightyear{2022}
\acmYear{2022}
\setcopyright{acmcopyright}
\acmConference[KDD '22] {Proceedings of the 28th ACM SIGKDD Conference on Knowledge Discovery and Data Mining}{August 14--18, 2022}{Washington, DC, USA.}
\acmBooktitle{Proceedings of the 28th ACM SIGKDD Conference on Knowledge Discovery and Data Mining (KDD '22), August 14--18, 2022, Washington, DC, USA}
\acmPrice{15.00}
\acmISBN{978-1-4503-9385-0/22/08}
\acmDOI{10.1145/3534678.3539453}

\begin{document}

\title{
Disentangled Ontology Embedding for Zero-shot Learning
}

\author{Yuxia Geng}
\email{gengyx@zju.edu.cn}
\affiliation{%
  \institution{College of Computer Science and Technology, 
  Zhejiang University}
  \city{Hangzhou}
  \country{China}
}

\author{Jiaoyan Chen}
\email{jiaoyan.chen@cs.ox.ac.uk}
\affiliation{%
  \institution{ Department of Computer Science,
  University of Oxford}
  \city{Oxford}
  \country{United Kingdom}}

\author{Wen Zhang}
\email{zhang.wen@zju.edu.cn}
\affiliation{%
  \institution{School of Software Technology, 
  Zhejiang University}
  \city{Ningbo}
  \country{China}
}
\author{Yajing Xu}
\email{yajingxu@zju.edu.cn}
\affiliation{%
  \institution{School of Software Technology, 
  Zhejiang University}
  \city{Ningbo}
  \country{China}
}

\author{Zhuo Chen}
\email{zhuo.chen@zju.edu.cn}
\affiliation{%
 \institution{College of Computer Science and Technology, 
  Zhejiang University}
  \city{Hangzhou}
  \country{China}
}

\author{Jeff Z. Pan}
\email{j.z.pan@ed.ac.uk}
\affiliation{%
 \institution{School of Informatics, The University of Edinburgh}
 \city{Edinburgh}
 \country{United Kingdom}}

\author{Yufeng Huang}
\email{huangyufeng@zju.edu.cn}
\affiliation{%
  \institution{School of Software Technology, 
  Zhejiang University}
  \city{Ningbo}
  \country{China}
}

\author{Feiyu Xiong}
\email{feiyu.xfy@alibaba-inc.com}
\affiliation{
  \institution{Alibaba Group}
  \city{Hangzhou}
  \country{China}
  }

\author{Huajun Chen}
\authornote{Corresponding author.}
\email{huajunsir@zju.edu.cn}
\affiliation{
\institution{College of Computer Science and Technology, Zhejiang University
\\
ZJU-Hangzhou Global Scientific and Technological Innovation Center
\\
Alibaba-Zhejiang University Joint Institute of Frontier Technologies
}
\city{}
  \country{}
}


\begin{abstract}
Knowledge Graph (KG) and its variant of ontology have been widely used for knowledge representation, and have shown to be quite effective in augmenting Zero-shot Learning (ZSL).
However, existing ZSL methods that utilize KGs all neglect the intrinsic complexity of inter-class relationships represented in KGs. One typical feature is that a class is often related to other classes in different semantic aspects.
In this paper, we focus on ontologies for augmenting ZSL, 
and propose to learn disentangled ontology embeddings guided by ontology properties to 
capture and utilize more fine-grained class relationships in different aspects.
We also contribute a new ZSL framework named \textbf{DOZSL}, which contains two new ZSL solutions based on generative models and graph propagation models, respectively, for effectively utilizing the disentangled ontology embeddings.
Extensive evaluations have been conducted on five benchmarks across zero-shot image classification (ZS-IMGC) and zero-shot KG completion (ZS-KGC).
DOZSL often achieves better performance than the state-of-the-art, and its components have been verified by ablation studies and case studies.
Our codes and datasets are available at \url{https://github.com/zjukg/DOZSL}.
\end{abstract}




\maketitle

\section{Introduction}

Zero-shot Learning (ZSL), which 
enables models to predict new classes that have no training samples (i.e., \textit{unseen classes}), has attracted a lot of research interests in many machine learning tasks, such as image classification \cite{xian2019survey,frome2013devise}, relation extraction \cite{li2020logic} and Knowledge Graph (KG) completion \cite{qin2020zsgan,wang2021structure}.
To handle these unseen classes, most existing ZSL  methods adopt a knowledge transfer strategy: transferring samples, sample features or model parameters from the classes that have training samples (i.e., seen classes) to these unseen classes, with the guidance of some auxiliary information which usually depicts the relationships between classes.
For example, in zero-shot image classification (ZS-IMGC), some studies utilize visual attributes of objects to transfer image features learned from seen classes to unseen classes and build classifiers for the later \cite{lampert2013attribute,xian2018feature}.
Other popular auxiliary information includes class's literal name \cite{frome2013devise}, textual descriptions \cite{qin2020zsgan,zhu2018noisy} and so on.

\begin{figure}
\centering
\includegraphics[width=0.49\textwidth]{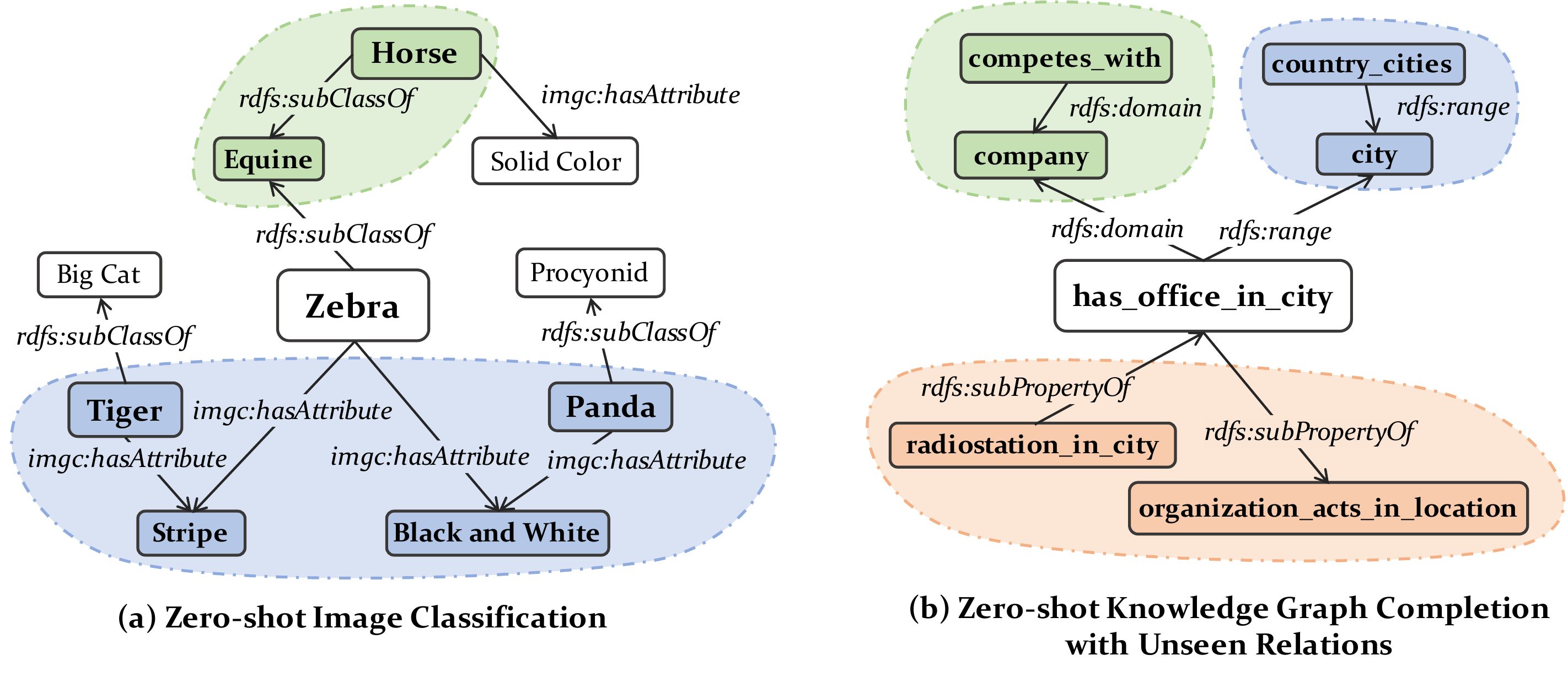}
\vspace{-5mm}
\caption{
(a) an ontology segment for zero-shot image classification where \textit{Zebra} is an unseen class while the other animals are seen classes; and (b) an ontological schema segment for zero-shot KG completion where \textit{has\_office\_in\_city} is an unseen relation while the other relations are seen.
The unseen class (or relation) connects itself to different seen classes (or relations) in different semantic aspects.
}
\label{fig:intro}
\vspace{-5mm}
\end{figure}

Recently, more and more studies leverage KG~\cite{Pan2016,hogan2021knowledge}, an increasingly popular solution for managing graph structured data, to represent complex auxiliary information for augmenting ZSL  \cite{chen2021knowledge}.
KGs that are composed of relational facts can model diverse relationships between classes.
For example, Wang et al. \cite{wang2018zero} incorporate class hierarchies from a lexical KG named WordNet \cite{miller1995wordnet}; works such as \cite{roy2020improving,geng2021benchmarking} explore common sense class knowledge from ConceptNet \cite{speer2017conceptnet}.
As a kind of KGs, ontologies, also known as ontological schemas when they act as parts of KGs for meta information, can represent more complex and logical inter-class relationships.
For example, Chen et al. \cite{chen2020ontology} use an ontology in OWL\footnote{Web Ontology Language (\url{https://www.w3.org/TR/owl-features/})} to express the compositionality of classes; Geng et al. \cite{geng2021ontozsl} define the domain and range constraints of KG relations using ontological schemas, as shown in Figure~\ref{fig:intro} (b).
In addition, ontologies are also able to represent and integrate traditional auxiliary information such as attributes and textual descriptions.
For example, as Figure~\ref{fig:intro} (a) shows, animal visual attributes with binary values can be represented in graph with the attributes transformed into entities.

To exploit these KGs, two ZSL paradigms have been widely
investigated.
One is a pipeline including two main steps.
Firstly, the KG is embedded, based on which the ZSL classes that are already aligned with KG entities are represented using vectors with their relationships kept in the vector space. 
Secondly, 
a compatibility function between the class vector and the sample input (or features) is learned.
It can either be a mapping function, which projects the sample input and the class vector into the same space such that a testing sample can be matched with an arbitrary class via e.g., Euclidean distance \cite{li2020logic,chen2020ontology,frome2013devise},
or a generative model, which generates labeled samples or features for unseen classes \cite{geng2021ontozsl,qin2020zsgan}.
The other paradigm is based on graph information propagation.
It often uses Graph Neural Networks (GNNs) to propagate classifier parameters or sample features from nodes of seen classes to nodes of unseen classes \cite{wang2018zero,kampffmeyer2019rethinking,chen2020zero}.
Methods of both paradigms, together with KGs, always lead to state-of-the-art performance on many ZSL tasks.

Nevertheless, existing methods of both paradigms still have big space for improvement.
In a real-world KG, an entity is often linked to other entities for knowledge of different aspects.
For example, Kobe Bryant is connected to NBA teams for his career knowledge, and connected to his daughters for family knowledge.
This also happens in those KGs (especially ontologies) used for augmenting ZSL.
As shown in Figure~\ref{fig:intro} (a), 
\textit{Zebra} is connected to \textit{Horse} via \textit{rdfs:subClassOf} for knowledge on taxonomy,
and connected to  \textit{Tiger} and \textit{Panda} via \textit{imgc:hasAttribute} for knowledge on visual characteristics.
Thus the vector representation of \textit{Zebra} should be closer to \textit{Horse} than \textit{Tiger} and \textit{Panda} considering the aspect of taxonomy, and be closer to  \textit{Tiger} and \textit{Panda} than \textit{Horse} considering the visual characteristics.
The existing KG-based ZSL methods all neglect this important KG characteristic on entanglement, which prevents them from 
capturing more fine-grained inter-class relationships in different aspects, and limits their performance.

In this work, we focused on augmenting ZSL by ontologies, proposed to investigate \textbf{D}isentangled \textbf{O}ntology embeddings and developed 
a general \textbf{ZSL} framework named \textbf{DOZSL}.
 DOZSL first learns multiple disentangled vector representations (embeddings) for each class according to its semantics of different aspects defined in an ontology, where a new disentangled embedding method with ontology property-aware neighborhood aggregation and triple scoring is proposed, 
 and then adopts an entangled ZSL learner, which builds upon a Generative Adversarial Network (GAN)-based generative model and a Graph Convolutional Network (GCN)-based graph propagation model, respectively, to incorporate these disentangled class representations.
To apply the generative model, we concatenate the disentangled representations; while to apply the propagation model, we generate one graph for semantics of one aspect with the disentangled representations.
%
We evaluate DOZSL with five datasets of zero-shot image classification (ZS-IMGC) and zero-shot KG completion (ZS-KGC).
See Figure~\ref{fig:intro} for segments of the ontology for one IMGC dataset and the ontological schema for a KG to complete.
%
In summary, our contributions are the following:
\begin{itemize}
\item To the best of our knowledge, this is among the first to investigate disentangled semantic embeddings for ZSL.
\item A property guided disentangled embedding method is developed for ontologies used in ZSL, and a general ZSL framework named DOZSL, which is able to support both generative models and propagation models, is proposed.
\item The work includes extensive evaluation,
where DOZSL often outperforms the baselines including the state-of-the-art methods on five datasets of two tasks, and the effectiveness of DOZSL's components is verified by ablation studies. 
\end{itemize}

\section{Preliminaries and Related Work}

\subsection{Zero-shot Learning}\label{zsl_formulation}

\subsubsection{Zero-shot Image Classification (ZS-IMGC)}\label{imgc_formulation}
ZSL has been thoroughly studied in Computer Vision for image classification with new classes whose images are not seen during training.
Formally, let $\mathcal{D}_{tr} = \{(x, y) | x \in \mathcal{X}_s, y \in \mathcal{Y}_s\}$ be the training set, where $x$ is the CNN features of a training image and $y$ is its class in $\mathcal{Y}_s$ which is a set  of seen classes, and $\mathcal{D}_{te} = \{(x, y) | x \in \mathcal{X}_u, y \in \mathcal{Y}_u\}$ be the testing set,
where $\mathcal{Y}_u$, the set of unseen classes, has no overlap with $\mathcal{Y}_{s}$.
Given $\mathcal{D}_{tr}$ and some auxiliary information $\mathcal{A}$ for describing the relationships between seen and unseen classes, ZS-IMGC aims to learn a classifier for each unseen class.
There are often two evaluation settings: standard ZSL which recognizes the testing samples in $\mathcal{X}_u$ by only searching in $\mathcal{Y}_u$ and generalized ZSL which recognizes the testing samples in $\mathcal{X}_s \cup \mathcal{X}_u$ by searching in $\mathcal{Y}_s \cup \mathcal{Y}_u$.

Widely used auxiliary information includes class attributes \cite{lampert2013attribute,xian2018feature,xu2020attribute}, textual information \cite{frome2013devise,zhu2018noisy} and KGs \cite{wang2018zero,roy2020improving,geng2021ontozsl,chen2021zero}.
To support ZSL, they are often embedded to generate one semantic vector for each class, such as binary/numerical attribute vectors, pre-trained word embeddings, learnable sentence embeddings, and KG embeddings.
Next, a compatibility function between the class vectors and the vector representations of samples is often learned to conduct knowledge transfer.
Mapping function is a typical practice, which maps the image features to the space of class vector \cite{frome2013devise,lampert2013attribute,chen2020ontology} or vice versa \cite{zhang2017learning} or to a shared common space \cite{fu2015zero}.
However, all of these mappings are trained by seen data, and thus have a strong bias towards seen classes during prediction, especially in generalized ZSL.
Recently, thanks to generative models such as GANs \cite{goodfellow2014generative}, several methods  \cite{xian2018feature,zhu2018noisy} have been proposed to synthesize samples (or features) for unseen classes conditioned on their class vectors.
This converts the ZSL problem to a standard supervised learning problem with the aforementioned bias issue alleviated.

Besides, to explicitly exploit the structural inter-class relationships that exist in a KG, some ZSL works explore a graph information propagation strategy.
In these works, classes are often aligned with KG entities, and a powerful GNN such as GCN \cite{kipf2016semi} is then trained to output a classifier
(i.e., a class-specific parameter vector) 
for each class, 
through which the classifiers of unseen classes are approximated by aggregating the classifiers of seen classes.
One typical work is by Wang et al. \cite{wang2018zero}, the subsequent works adopt similar ideas but vary in optimizing the graph propagation \cite{kampffmeyer2019rethinking,geng2020explainable}.
Especially, some of them consider the multiple types of relations in the KGs by developing multi-relational GCN \cite{chen2020zero}, or spliting the multi-relation KGs into multiple single-relation graphs and applying several parameter-shared GCNs to propagate features \cite{wang2021zero}.


\subsubsection{Zero-shot KG Completion (ZS-KGC)}\label{kgc_formulation}

In this task, a KG composed of relational facts is to be completed. 
It is denoted as $\mathcal{G} =\{ \mathcal{E}, \mathcal{R}, \mathcal{T}\}$, where $\mathcal{E}$ is a set of entities, $\mathcal{R}$ is a set of relations, and $\mathcal{T} = \{(h, r, t)| h, t \in \mathcal{E}; r \in \mathcal{R}\}$ is a set of relational facts in form of RDF triple.
The completion is to predict a missing but plausible triple with two of $h,r,t$ given. 
Typical KGC methods first embed entities and relations into vector spaces (i.e., $x_h, x_r$ and $x_t$) and conduct vector computations to discover missing triples.
The embeddings are trained by existing triples and assume all testing entities and relations are available at training time.
ZS-KGC is thus proposed to predict for unseen entities or relations that are newly added during testing and have no associated training triples.

Some ZS-KGC approaches devote to dealing with unseen entities by utilizing the auxiliary connections with seen entities \cite{wang2019logic}, introducing their textual descriptions \cite{wang2021structure},
or learning entity-independent graph representations so that naturally generalizing to unseen entities \cite{teru2019inductive,MorsE}.
In contrast, the works for unseen relations are relatively underexplored. Both Qin et al. \cite{qin2020zsgan} and Geng et al. \cite{geng2021ontozsl}  leverage GANs to synthesize valid embeddings for unseen relations conditioned on their auxiliary information which are textual descriptions and ontological schemas, respectively.

In this study, we target at unseen relations.
Two disjoint relation sets: the seen relation set $\mathcal{R}_s$ and the unseen relation set $\mathcal{R}_u$ are set.
The triple set $\mathcal{T}_s = \{(h, r_s, t)| h, t \in \mathcal{E}; r_s \in \mathcal{R}_s\}$ is collected for training, and $\mathcal{T}_u = \{(h, r_u, t)| h, t \in \mathcal{E}; r_u \in \mathcal{R}_u\}$ is collected to evaluate the completion of the triples of unseen relations.
A closed set of entities is considered following previous works, i.e., each entity that appears in the testing set has appeared during training.


\subsection{Ontology}\label{ontology_formulation}
Ontology is famous for representing and exchanging general or domain knowledge, often with hierarchical concepts as the backbone and properties for
describing semantic relationships \cite{horrocks2008ontologies}.
In this study, we use a simple form of ontology, namely in RDF Schema (RDFS)\footnote{\url{https://www.w3.org/TR/rdf-schema/}}, while those more complicated OWL ontologies can be transformed into RDFS ones following some criteria.
An ontology can be used as a schema of a KG, defining entity types, relations and so on.
Accordingly, we represent an ontology as $\mathcal{O}=\{\mathcal{C}, \mathcal{P}, \mathcal{T}_o\}$, where $\mathcal{C}$ is the set of concepts (a.k.a. types), $\mathcal{P}$ is the set of properties, and $\mathcal{T}_o=\mathcal{C} \times \mathcal{P} \times \mathcal{C}$ is the set of triples.
To serve as auxiliary information for ZSL, an ontology models the relevant domain knowledge of a given ZSL task.
For example, in IMGC, concepts are used to represent image classes and image attributes; in KGC, ontology triples can be used to define    
domains (i.e., head entity types) and ranges (i.e., tail entity types) of KG relations.
Note we sometimes also call concept as concept node in introducing ontology embedding.

Ontology properties can be either built-in properties of RDFS, such as \textit{rdfs:subClassOf} and \textit{rdfs:subPropertyOf}, or user defined for a specific task, 
such as \textit{imgc:hasAtrtibute}. 
Figure~\ref{fig:intro} shows two ontology segments for ZS-IMGC and ZS-KGC.
The triple (\textit{Zebra, imgc:hasAttribute, Stripe}) means that an animal class \textit{Zebra} has an attribute \textit{Stripe} in decoration,
while the triple (\textit{radiostation\_in\_city, rdfs:subPropertyOf, has\_office\_in\_city}) means that the KG relation \textit{radiostation\_in\_city} is a subrelation of  \textit{has\_office\_in\_city}.  
It is worth mentioning that properties are also often defined with hierarchies, as the concepts. One general property is often defined for semantics of one aspect, and then more sub-properties are defined for more fine-grained semantics. Thus we can often easily find out relevant properties for different semantic aspects of an ontology by simple visualization of the property hierarchies.

In our ZS-KGC case study, we adopt ontologies developed in \cite{geng2021ontozsl} as the auxiliary information for completing relational facts of their corresponding KGs in the zero-shot setting, where KG relations are modeled as ontology concepts and their meta-relationships are modeled by ontology properties.
Our DOZSL framework contains a disentangled ontology encoder to learn disentangled representations for all concept nodes in an ontology, through which the fined-grained inter-concept relationships can be figured out and well utilized in downstream zero-shot learning and prediction steps.

\subsection{Disentangled Representation Learning}
The goal of disentangled representation learning is to learn embedding including various separate components behind the data.
In the field of the graph, DisenGCN \cite{ma2019disentangled} is the first work tending to learn disentangled node representations, which uses a neighborhood routing mechanism to identify the latent factor that may have caused the link from a given node to one of its neighbors.
However, it mainly focuses on homogeneous graphs with a single relation type.
To process graphs with more diverse relation types,
DisenE \cite{kou2020disene} and DisenKGAT \cite{wu2021disenkgat}, which leverage an attention mechanism and a dynamic assignment mechanism, respectively, disentangle the entity embeddings according to the relations in a KG.
Different from these works, we propose to learn disentangled ontology embeddings in terms of the characteristics of the ontology used for ZSL and develop a novel disentanglement mechanism which is guided by the properties in an ontology.

There are also some works that explore the disentangled representation learning in ZSL \cite{xu2020attribute,li2021generalized}.
However, they all focus on disentangling the representations of samples such as the image features learned by CNNs, none of them have  taken into account the impact of learning disentangled auxiliary information representations, especially when richer but complex auxiliary information are introduced.
In contrast, our work made the first attempt.


\begin{figure*}
\centering
\includegraphics[height=5cm]{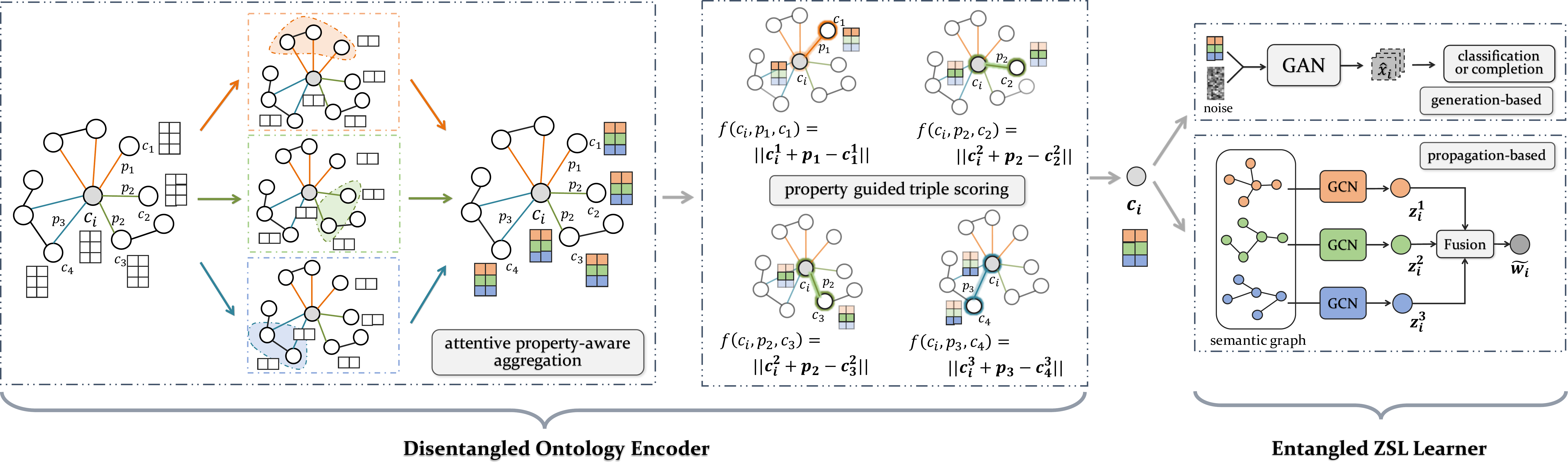}
\vspace{-2.5mm}
\caption{Illustration of DOZSL with $K=3$. Different color means different semantic aspects.}
\label{fig:framework}
\vspace{-3mm}
\end{figure*}

\section{Methodology}

As shown in Figure~\ref{fig:framework}, \textbf{DOZSL} includes two core modules: \textbf{Disentangled Ontology Encoder} learning disentangled ontology embeddings, and
\textbf{Entangled ZSL Learner} utilizing the embeddings for generation-based and propagation-based ZSL methods.

\subsection{Disentangled Ontology Encoder}\label{sec:doe}


In DOZSL, the embedding of each concept node $c$ is disentangled into multiple distinct components as $\bm{c} = [\bm{c}^1, \bm{c}^2, ..., \bm{c}^K]$, where $K$ is the component numbers,  $\bm{c}^k \in \mathbb{R}^{\frac{d}{K}}$ represents the $k$-th component encoding semantics of one aspect of $c$ and $d$ is the embedding size.

To learn disentangled embedding for each concept, we first aggregate information from its graph neighborhoods that characterize it. 
In the aggregation of each component for a concept, only a subset of neighbors actually carries valuable information since each component represents a specific semantic aspect. 
To identify the aspect-specific subset, we follow the attention-based neighborhood routing strategy in previous works \cite{ma2019disentangled,wu2021disenkgat}.
Also, considering the various relation types in the ontologies, we propose a \textbf{property-aware attention mechanism}.
Specifically, for the $k$-th aspect, the attention value of one neighbor $c_j$ of concept $c_i$ is computed by the similarity of the $k$-th component embeddings of $c_j$ and $c_i$ in the subspace of their connection property $p$ following the assumption that when a neighbor contributes more to $c_i$ in the aggregation, their property-aware representations are more similar, formally:
\begin{align}\label{eq:attention}
\alpha_{(c_i,p,c_j)}^{k,l} &= softmax((h_{i,k,p}^l)^T \cdot h_{j,k,p}^l)
    \nonumber \\
&= \frac{exp((h_{i,k,p}^l)^T \cdot h_{j,k,p}^l)}{\sum_{(c_{j'}, p')\in \mathcal{N}(i)} exp((h_{i,k,p'}^l)^T \cdot h_{j',k,p'}^l)} \\
h_{i,k,p}^l &= h_{i,k}^l \circ W_p, \;\;\; h_{j,k,p}^l = h_{j,k}^l \circ W_p 
\end{align}  
where $l\in \{0, 1, ..., L-1\}$ with $L$ as the number of aggregation layers. 
$h_{i,k,p}^l$ is
the $k$-th component embedding of $c_i$ w.r.t. property $p$ in the $l$-th aggregation layer, $\circ$ denotes the Hadamard product, and $W_p$ is a learnable projection matrix of $p$ for projecting $c_i$'s $k$-th component embedding $h_{i,k}^l$ into the property specific subspace.
$\mathcal{N}(i) = \{ (c_{j'}, p') | (c_i, p', c_{j'}) \in \mathcal{T}_o \} \cap \{ (c_i, p_s)\}$ is the set of pairs of neighboring concept nodes and properties of $c_i$, which also includes $c_i$ itself with a special self-connection property $p_s$. $\mathcal{T}_o$ is the ontology triple set. 
A dot-product similarity is adopted here.

With attention values, we separately aggregate the neighborhood information for representing each component and also update the property embedding after each aggregation as:
\begin{equation}\label{RGAT}
h_{i,k}^{l+1} = \sigma(\sum_{(c_j,p)\in \mathcal{N}(i)} \alpha_{(c_i,p,c_j)}^{k,l} \phi (h_{j,k}^l, h_p^l, W_p)), \;\;
h_p^{l+1} =  h_p^{l}\cdot \Theta_p^l
\end{equation}
where $h_p^l$ is the embedding of property $p$ in the $l$-th layer.
$\Theta_p^l$ is the layer-specific linear transformation matrix for $p$.
$\phi$ is a combination operator for fusing the information of neighboring concept nodes and property edges.
Here, we refer to CompGCN \cite{vashishth2020composition} to implement it via e.g. vector multiplication.
$h_{i,k}^0$ is randomly initialized, and
$h_{i,k}^L$ is outputted at last layer which has encoded the neighborhood information specific to aspect $k$. We make $\bm{c}_i^k  = h_{i,k}^L$ for simplicity.

To further improve the disentanglement, 
we propose to refine the semantics of each disentangled component embedding of concepts according to their associated properties.
It is inspired by the characteristic of knowledge in ontologies, i.e., ontology properties are often represented with hierarchies, thus one general property can always be selected for representing one distinct semantic aspect of a concept; for example, the properties \textit{imgc:hasAttribute} and \textit{rdfs:subClassOf} in Figure~\ref{fig:intro} represent the semantics on animal visual characteristics and taxonomy, respectively.

To achieve this goal, we \textit{(i)} select a set of properties for aspects of the semantics of the concepts to encode (e.g., \textit{imgc:hasAttributes} for visual characteristics in the ontology for IMGC) and set the number of disentangled components to be the number of selected properties, 
and \textit{(ii)} design a \textbf{property guided triple scoring mechanism} extracting property-specific components to constitute a valid ontology triple.
Specifically, 
for an ontology triple ($c_i, p_k, c_j$), we extract the $k$-th components of $\bm{c}_i$ and $\bm{c}_j$ with respective to property $p_k$, and 
leverage the score function on KG embedding methods to calculate the triple score with the extracted components.
In this way, we accurately endow each component embedding with a specific semantic meaning w.r.t properties.
Here, the score function of TransE \cite{bordes2013translating} is adopted to compute the triple score as: 
\begin{equation}\label{eq:score}
q_{(c_i, p_k, c_j)} = f(-||\bm{c}_i^k + \bm{p}_k -  \bm{c}_j^k||)
\end{equation}
where $\bm{c}_i^k$ and $\bm{c}_j^k$ denote the extracted component embeddings of concepts $c_i$ and $c_j$ respectively, and $\bm{p}_k$ represents the embedding of property $p_k$.
$f$ is the logistic sigmoid function.
A higher score indicates a stronger relatedness between $\bm{c}_i^k$, $\bm{p}_k$ and $\bm{c}_j^k$.
Finally, we use the standard cross entropy with label smoothing to train the whole disentangled ontology encoder as:
\begin{equation}
\begin{aligned}
\mathcal{L} = -\frac{1}{B}\frac{1}{|\mathcal{C}|} \sum_{(c_i, p_k) \in batch} \sum_{n} (t_n \cdot log(q_{(c_i, p_k, c_j^n)}) + \\ (1-t_n) \cdot log(1- q_{(c_i,p_k,c_j^n)}))
\end{aligned}
\end{equation}
where $B$ is the batch size, $\mathcal{C}$ is the concept node set of the ontology, $t_n$ is the label of the given query $(c_i, p_k)$, whose value is $1$ when the triple $(c_i, p_k, c_j^n)$ holds and $0$ otherwise.

\subsection{Entangled ZSL Learner}
With the disentangled ontology embeddings, we next show how to utilize them for ZSL.
Specifically, we develop two kinds of methods.
In consideration of the effectiveness of GANs in learning the compatibility between class vectors and their samples, the first method is generation-based leveraging GANs to generate discriminative samples for classes (each of which corresponds to an ontology concept).
The other is propagation-based propagating features among  classes based on the disentangled graphs generated from the original ontology.


\subsubsection{Generation-based}\label{sec:gb}
We first get the embedding of each class by concatenating all $K$ component embeddings of its corresponding ontology concept (i.e., $\bm{c}_i = [\bm{c}_i^1, \bm{c}_i^2, ..., \bm{c}_i^K]$) ,
and then adopt a typical scheme of GAN for feature generation.
Specifically, the GAN consists of three networks: a generator $G$ synthesizing sample features for a class from random noises conditioned on its embedding; a feature extractor $E$ providing the real sample features; and a discriminator $D$ distinguishing the generated features from the real ones.
We generate sample features instead of raw samples for both higher accuracy and efficiency, as in many works \cite{xian2018feature,geng2021ontozsl,qin2020zsgan}.

Formally,  for a class $c_i$, the generator $G$ takes as input 
its embedding and 
a random noise vector $z$ sampled from Normal distribution, 
and generates its features: $\hat{x} = G(z, \bm{c}_i)$.
The loss of $G$ is defined as:
\begin{equation}
	\mathcal{L}_G = - \mathbb{E}[D(\hat{x})] + \lambda_1 \mathcal{L}_{cls} (\hat{x}) + \lambda_2 \mathcal{L}_{R}
\end{equation}
where the first term is the Wasserstein loss, the second term 
is a supervised classification loss for classifying the synthesized features,
and 
the third is
for regularizing the mean of generated features of each class to be the mean of its real features.
The latter two  both encourage the generated features to have more inter-class discrimination.
$\lambda_1$ and $\lambda_2$ are the corresponding weight coefficients.

The discriminator $D$ takes as input the synthesized features $\hat{x}$ from $G$ and the real features $x$ from $E$. Its loss is defined as:
\begin{equation}
	\mathcal{L}_D = \mathbb{E}[D(x, \bm{c}_i)] - \mathbb{E}[D(\hat{x})] - \beta \mathbb{E}[(|| \bigtriangledown_{\tilde{x}} D(\tilde{x}) ||_p -1)^2]
\end{equation}
where the first two terms approximate the Wasserstein distance of the distributions of $x$ and $\hat{x}$.
The last term is the gradient penalty to enforce the gradient of $D$ to have unit norm 
in which $\tilde{x} = \varepsilon x + (1-\varepsilon) \hat{x}$ with $\varepsilon \sim U(0,1)$. 
$\beta$ is the weight coefficient.

In view of the different data form in different ZSL tasks, we adopt different feature extractor $E$.
For ZS-IMGC, we employ ResNet101 \cite{he2016resnet} to extract the features of images following previous works \cite{xian2019survey}; and for ZS-KGC, we follow \cite{qin2020zsgan,geng2021ontozsl} to learn cluster-structured features for KG relations.
In general, $E$ is trained in advance with only samples of seen classes, 
and is fixed during adversarial training.
Also, our framework is compatible to different feature extractors.

With well trained GAN,
we use generator $G$ to synthesize features and train task-specific prediction models for unseen classes.
In ZS-IMGC, we train a softmax classifier for each unseen class to classify its testing images;
in ZS-KGC, a testing triple is completed by calculating the similarity between the generated embedding of the relation $r$ and the joint embedding of the entity pair $(h,t)$. 

\subsubsection{Propagation-based}
With disentangled concept embeddings, 
more fine-grained relatedness between concepts could be utilized.
Therefore, as shown in Figure~\ref{fig:framework}, we generate one semantic graph for each component, where nodes correspond to the classes (relations in KGC) in the dataset and  edges are generated by calculating the cosine similarity between the component embeddings of two class nodes, and conduct graph propagation on it to transfer features between classes under each semantic aspect.
The initialized node features are  the class's component embedding.
Formally, we represent the $k$-th semantic graph as $\mathbf{G}_k(A_k, S_k)$, where $S_k \in \mathbb{R}^{m \times \frac{d}{K}}$ is the input feature matrix of graph nodes, and $A_k \in \mathbb{R}^{m \times m}$ is the graph adjacency matrix indicating the connections among $m$ classes defined as below, $\tau$ denotes the similarity threshold.
\begin{equation}
A_k (i,j) = 
\left\{
\begin{array}{cc}
    1 & \text{if} \quad sim(\bm{c}_i^k, \bm{c}_j^k) > \tau
    \\
    0 & \text{otherwise}
\end{array}
\right.
\end{equation}

Since $\mathbf{G}_k$ is a graph with one single relation, we use GCN for feature propagation. Each graph convolutional layer performs as: 
\begin{equation}
    H_k^{l+1} = \sigma (\widehat{A}_k H_k^{l} \Phi_k^{l})
\end{equation}
where 
$\widehat{A}_k$ is the normalized adjacent matrix, and $\Phi_k^{l}$ is a layer-specific weight matrix shared among all semantic graphs.
$H_k^{0}=S_k$.

For each semantic graph, the GCN outputs a set of  node embeddings $Z_k \in \mathbb{R}^{m \times F}$, through which we can obtain a set of classifiers $\widetilde{\mathcal{W}}$ for all $m$ classes as: $\widetilde{\mathcal{W}} = \varphi (Z_1, Z_2, ..., Z_K)$, where $\varphi$ is a fusion function.
In our experiments, we implement $\varphi$ by averaging: $\widetilde{\mathcal{W}}=\frac{1}{K}\sum_{k} Z_k$, or linear transformation:
$\widetilde{\mathcal{W}} = W_1([Z_1; Z_2; ...; Z_K])$ where 
$W_1 \in \mathbb{R}^{KF\times F}$ is a trainable transformation matrix.
Then, following \cite{wang2018zero,kampffmeyer2019rethinking,wang2021zero}, we compute the Mean Square Error between the fused classifiers and the ground-truth classifiers as loss function:
\begin{equation}
    \mathcal{L}_{GCN} = \frac{1}{|\widetilde{\mathcal{W}_s}|} \sum_{\widetilde{w} \in \widetilde{\mathcal{W}_s}} (\widetilde{w} - gt(\widetilde{w}))^2
\end{equation}
where $\widetilde{\mathcal{W}_s} \subset \widetilde{\mathcal{W}}$ is the set of classifiers of the seen classes,
$gt(\widetilde{w})$ denotes the corresponding ground-truth.
Different from the traditional classifier which is a network trained using labeled samples, the classifier here is actually a real-valued vector that represents the class-specific features, and is obtained by averaging the features of all the training samples of one class in our paper.
The sample features are also extracted via the  feature extractor $E$ mentioned in Section \ref{sec:gb}. 
By using these ground-truth seen classifiers to supervise the training of GCNs, classifiers of the unseen classes can be learned by aggregation.
During prediction, for an input testing sample,
we first extract its features using the same feature extractors, and then perform classification or completion by calculating the similarity between the learned classifiers and the extracted features.

\section{Evaluation}

\subsection{Experiment Settings}

\subsubsection{Datasets and Ontologies}
For ZS-IMGC, we use a popular benchmark named Animals with Attributes (AwA) \cite{xian2019survey} and two benchmarks ImNet-A and ImNet-O extracted from ImageNet by Geng et al. \cite{geng2021ontozsl}.
AwA  is for coarse-grained animal image classification wth $50$ classes and $37,322$ images.
ImNet-A is for more fine-grained animal image classification and ImNet-O is for fine-grained general object classification. 
The classes are split into a seen set and an unseen set, following \cite{xian2019survey}.
For ZS-KGC, we use two KGs provided in \cite{qin2020zsgan} for completion, i.e., NELL-ZS and Wiki-ZS extracted from NELL and Wikidata\footnote{NELL (\url{http://rtw.ml.cmu.edu/rtw/}) and Wikidata (\url{https://www.wikidata.org/})}, respectively.
In each KG, the relations are split into a training set with seen relations, a validation set and a testing set with unseen relations, following \cite{qin2020zsgan}.
Accordingly, their associated triples compose a training set, a validation set and a testing set.
It is ensured that all entities are seen.

Each dataset has an ontology as its auxiliary information.
We use the ontologies developed in \cite{geng2021ontozsl} and take the latest version released in \cite{geng2021benchmarking}.
For ZS-IMGC, the ontologies contain class hierarchies (taxonomies), class visual attributes and attribute hierarchies.
In our property guided disentangled embedding, we select two general properties: \textit{rdfs:subClassOf} for semantic aspect on taxonomy, and \textit{imgc:hasAttribute} for semantic aspect on visual characteristics.
For ZS-KGC, the ontologies contain type constraints of the head and tail entities of relations, represented by properties \textit{rdfs:domain} and \textit{rdfs:range}, relation hierarchies represented by property \textit{rdfs:subProperty}, and type hierarchies represented by property \textit{rdfs:subClassOf}.
These four properties are selected as general properties used in ontology encoder.
See Table~\ref{tab:datasets} for detailed statistics.

\begin{table}
\small
\centering
\caption{\small Statistics of benchmarks in two ZSL tasks and their ontologies. Trip./Conp./Prop. in the column of \# Ontologies denotes the number of triples/concepts/properties. S/U denotes seen/unseen classes.
Tr/V/Te is short for  training/validation/testing.}\label{tab:datasets}
\vspace{-2.5mm}
\begin{tabular}{c|c|p{0.5cm}<{\centering}p{0.8cm}<{\centering}p{1.4cm}<{\centering}|p{1.8cm}<{\centering}}
\hline
\multirow{3}{*}{\textbf{Datasets}} 
&
\multirow{2}{*}{\textbf{\#Classes}} & \multicolumn{3}{c|}{ \textbf{\#Images}} 
& \multirow{2}{*}{\textbf{\# Ontologies}}
\\
& & 
& Training & Testing & \\ 
& \multicolumn{1}{c|}{Total(S/U)}
&Total & S/U 
& \multicolumn{1}{c|}{S/U}
& Trip./Conp./Prop.
\\\hline
 AwA 
& 50(40/10) 
& 37,322  & 23,527/0 & 5,882/7,913
& 1,759 / 202 / 2 
\\
ImNet-A 
& 80(28/52)
& 77,323 & 36,400/0 & 1,400/39,523
& 545 / 214 / 2
\\
ImNet-O 
& 35(10/25)
& 39,361  & 12,907/0
& \ \ \ 500/25,954
&220 / 111 / 2
\\
\hline
\end{tabular}
\begin{flushleft}
\end{flushleft}
\begin{tabular}{l|p{0.6cm}<{\centering}p{0.6cm}<{\centering}cc}
\hline
\multirow{2}{*}{\bf Datasets} 
& \multicolumn{1}{c}{\multirow{2}{*}{\bf \#Entity }} 
& \multicolumn{1}{c}{\multirow{2}{*}{\bf \#Triples }} 
& \multicolumn{1}{c}{\bf \#Relations}
& \multicolumn{1}{c}{\bf \# Ontologies} \\
& &  & \multicolumn{1}{c}{ Tr/V/Te} 
 & \multicolumn{1}{c}{Trip./Conp./Prop.} 
\\\hline
NELL-ZS  
& 65,567 & 188,392  & 139/10/32 
& 3,055 / 1,186 / 4  \\
Wiki-ZS     
& 605,812 &  724,967 & 469/20/48
& 4,821 / 1,904 / 4
\\
\hline
\end{tabular}

\vspace{-0.3cm}
\end{table}

\subsubsection{Variants of DOZSL and Baselines}

In disentangled ontology encoder, we compare two settings for component embeddings that are fed to score triple (Eq. \eqref{eq:score}): aggregating neighborhood information (Eq. \eqref{eq:attention} and \eqref{RGAT}), and randomly initializing component embeddings without neighborhood aggregation.
This leads to two DOZSL variants.
Meanwhile, they can be combined with two downstream ZSL methods: generation-based with GAN and propagation-based with GCN.
Thus we have four DOZSL variants and denote them as ``DOZSL(X+Y)'', where X can be AGG (neighborhood aggregation) and RD (random initialization), Y can be GAN and GCN.


The baselines include those generation-based and propagation-based ZSL methods that often achieve state-of-the-art performance on many ZSL datasets.
\textbf{OntoZSL} \cite{geng2021ontozsl} is a generation-based method that uses GANs to synthesize samples, where  we take TransE as its ontology encoder for a fair comparison.
\textbf{DGP} \cite{kampffmeyer2019rethinking} is a propagation-based method using a two-layers GCN which only supports single-relation graphs.
To deal with the multi-relation ontology graph, we take the method proposed in \cite{wang2021zero} as a baseline.
Meanwhile, two relation-aware GNNs, RGCN \cite{schlichtkrull2018rgcn} and CompGCN \cite{vashishth2020composition}, are also used to implement another two propagation-based ZSL baselines.
%
We also consider different disentangled and non-disentangled semantic embedding methods for more baselines.
For non-disentangled embedding, we choose classical \textbf{TransE}, and \textbf{RGAT} which also performs attentive relation-aware graph  aggregation.
For disentangled embedding, we choose two state-of-the-art methods \textbf{DisenE} \cite{kou2020disene} and \textbf{DisenKGAT} \cite{wu2021disenkgat}.
These embedding methods can also be combined with GAN-based and GCN-based ZSL learners as in DOZSL, leading to baselines such as ``DisenKGAT+GAN''.
Note ``TransE+GAN'' is equivalent to OntoZSL.



\subsubsection{Evaluation Metrics}
For ZS-IMGC, we report macro accuracy following \cite{xian2019survey}, where accuracy of each class is first calculated with its testing images, and the accuraccies of all testing classes are then averaged.
For standard ZSL testing, we compute accuracy on all unseen classes, denoted as $acc$; while for generalized ZSL testing, we first calculate accuracy for all the seen classes and all the unseen classes separately, denoted as $acc_s$ and $acc_u$, respectively, and then report a harmonic mean $H = (2 \times acc_s \times acc_u)/(acc_s + acc_u)$.


\begin{table*}[]
\small
\caption{$Acc$uracy and $H$ ($\%$) of ZS-IMGC on AwA, ImNet-A and ImNet-O. $MRR$ and $hit@k$ ($\%$) of ZS-KGC on NELL-ZS and Wiki-ZS. The best results in a method category (resp. in the whole column) are in bold (resp. underlined).
TransE+GAN equals OntoZSL.
}
\label{tab:overall_results}
\vspace{-0.25cm}
\setlength\tabcolsep{3pt}
\begin{tabular}{c|l|p{0.7cm}<{\centering}p{0.7cm}<{\centering}|p{0.7cm}<{\centering}p{0.7cm}<{\centering}|p{0.7cm}<{\centering}p{0.7cm}<{\centering}|c|p{0.7cm}<{\centering}p{0.7cm}<{\centering}p{0.7cm}<{\centering}p{0.7cm}<{\centering}|p{0.7cm}<{\centering}p{0.7cm}<{\centering}p{0.7cm}<{\centering}p{0.7cm}<{\centering}}
\hline
\multicolumn{1}{c|}{\multirow{2}{*}{\bf Category}} & \multicolumn{1}{c|}{\multirow{2}{*}{\bf Methods}}  & \multicolumn{2}{c}{\bf AwA}   & \multicolumn{2}{c}{\bf ImNet-A} & \multicolumn{2}{c|}{\bf ImNet-O} &
&
\multicolumn{4}{c}{\bf NELL-ZS} & \multicolumn{4}{c}{\bf Wiki-ZS}
\\
& & $acc$ & $H$ & $acc$ & $H$ & $acc$ & $H$ & & $hit@10$ & $hit@5$ & $hit@1$ & $MRR$ & $hit@10$ & $hit@5$ & $hit@1$ & $MRR$ \\
\cline{1-8}\cline{10-17}
\multirow{6}{*}{Generation}
& \ TransE+GAN
& 58.28 & 54.77
& 39.44 & 31.61  
& 31.93 & 27.82 
&
& 34.9 & 29.0 & 15.6 & 22.5 
& 27.6 & 22.3 & 13.6 & 18.7
\\
& \ RGAT+GAN
& 63.95 &  57.52
& 39.20 &  31.17  
& 35.13 & 27.68 
&
& 34.8 & 28.7 & 16.2 & 22.7
& 27.9 & 22.5 & \underline{\textbf{14.2}} & \underline{\textbf{19.1}}
\\
\cline{2-8}\cline{10-17}
& \ DisenE+GAN
& 59.40 & 49.47 & 33.60 & 29.96
& 31.62 & 26.69 
&
& 34.8 & 29.1 & 15.3  & 22.2 
& \textbf{28.0} & 22.7 & 13.8 & 18.9 
\\
& \ DisenKGAT+GAN
& 61.81& 54.41
& 35.90 & 31.09 
& 34.94 & 27.33  
&
& 35.9 & 29.5 & 15.7 & 22.9
& 27.6 & 22.4 & 13.8 & 18.8
\\
\cline{2-8}\cline{10-17}
& \ DOZSL(RD+GAN)
& 52.35 & 46.91
& 37.12 & 30.18 
& 34.48 &  28.57
&
& \textbf{36.4} & \textbf{29.9} & \textbf{16.5}  & \textbf{23.4}
& 27.9 & \textbf{22.7} & 14.0 & 19.0
\\
& \ DOZSL(AGG+GAN)
& \underline{\textbf{66.36}} & \underline{\textbf{57.62}}
& \underline{\textbf{40.26}} & \textbf{32.82}  
& \textbf{36.00} & \textbf{28.74}  
&
& 36.2 & 29.5 & 16.1 & 23.0
& 27.7 & 22.7 & 13.3 & 18.6
\\
\cline{1-8}\cline{10-17}
\multirow{10}{*}{Propagation}
& \ DGP
& 59.03 & 28.97
& 35.72 & 29.98 
& 34.89 &  29.76  
&
& 36.2 & 29.5 & 16.1 & 23.0
& 27.7 & 22.7 & 13.3 & 18.6
\\
& \ Wang et al. \cite{wang2021zero} 
& 43.81 & 42.13 
& 34.33 & 21.95   
& 32.73 & 26.86
&
& 35.8 & 29.6 & 15.7 & 22.8 
& 26.8 & 21.9 & 13.5 & 18.3
\\
& \ RGCN-ZSL 
& 44.90 & 24.95 
& 37.36 & \underline{\textbf{33.01}}
& 31.19 & 23.39 
&
& 37.4 & 30.7 & \underline{\textbf{17.0}} & \underline{\textbf{24.1}}
& \underline{\textbf{28.5}} & \underline{\textbf{23.2}} & 13.7 & \underline{\textbf{19.1}}
\\
& \ CompGCN-ZSL     
& 53.46 & 29.33 
& 38.34 & 29.01 
& 28.95  & 27.35 
&
& 36.0 & 29.7 & 16.4 & 23.2  
& 28.0 & 22.7 & 13.5 & 18.8  
\\
\cline{2-8}\cline{10-17}
& \ TransE+GCN
& 63.56 &  36.15
& 36.69 & 22.12 
& 33.16 & 24.72
&
& 35.8 & 29.8 & 16.0 & 22.9
& 26.6 & 21.5 & 13.6 & 18.3
\\
& \ RGAT+GCN
& 58.83 & 37.35
& 37.53 & 31.27  
& 35.47 & 28.49
&
& 36.1 & 29.8 & 16.0 & 22.9
& 26.6 & 21.6 & 13.7 & 18.3 
\\
\cline{2-8}\cline{10-17}
& \ DisenE+GCN
& 58.34 & 50.86 
& 32.56 & 27.76 
& 32.02 & 26.33
&
& 35.5 & 29.7 & 15.6 & 22.7
& 26.7 & 21.7 & 13.7 & 18.3 
\\
& \ DisenKGAT+GCN
& 61.24 & 37.43
& 37.55 & 32.27
& 35.92  & 29.50
&
& 35.7 & 29.5 & 16.1 & 23.0
& 27.5 & 22.1 & 13.8 & 18.6
\\
\cline{2-8}\cline{10-17}
& \ DOZSL(RD+GCN)
& 62.79 & \textbf{52.74}
& 36.01 & 30.29  
& 33.66 & 31.19 
&
& \underline{\textbf{38.0}} & \underline{\textbf{31.2}} & 16.5 & 23.9
& 26.7 & 21.9 & \textbf{13.8} & 18.5 
\\
& \ DOZSL(AGG+GCN)
& \textbf{63.88} & 44.52
& \textbf{38.69} & 32.12  
& \underline{\textbf{37.42}} & \underline{\textbf{31.77}}
&
& 36.2 & 29.3 & 16.2 & 23.0
& 27.5 & 22.4  & 13.6 & 18.7
\\
\hline
\end{tabular}
\end{table*}

Our ZS-KGC task is to predict the tail entity $t$ given a head entity $h$ and an unseen relation $r_u$.
Thus for the input of a testing triple ($h, r_u$), we rank a set of candidate entities
according to their predicted scores of being the tail entity, and see the rank of the ground truth tail entity --- the smaller rank, the better performance. 
As in most KGC works, we report Mean Reciprocal Ranking ($MRR$) and $hit@k$ (i.e., the ratio of testing samples whose ground truths are ranked in the top-$k$ position).
$k$ is set to $1, 5, 10$.
%
Different from ZS-IMGC where predicting the class label of an image tends to be confused by other classes, the prediction for a seen relation in ZS-KGC is relatively independent of the prediction for an unseen relation. Thus the generalized ZSL testing setting in ZS-KGC, which is a simple addition of normal KGC, is not considered in our paper.

\vspace{-0.2cm}
\subsection{Main Results}

\subsubsection{ZS-IMGC}\label{imgc_results_report}

The results are reported based on these settings. 
For ontology encoder, we set the component embedding size and the property embedding size to $100$.
$K$ is set to $2$ (corresponding to \textit{rdfs:subClassOf} and \textit{imgc:hasAttribute}) for all DOZSL(RD) variants, but to $5$ for all DOZSL(AGG) variants since two reverse properties and a self-connection property are added during aggregation.
The initial learning rate is set to $0.001$. The number of the aggregation layer for DOZSL(AGG) variants is set to $1$.

For ZSL learner, we employ ResNet101 to extract $2,048$-dimensional image features.
It is ensured that unseen classes of all the three datasets have never appeared in training ResNet101.
Regarding GAN, the generator and discriminator both consist of two fully connected layers with $4,096$ hidden units;
their learning rates are both set to $0.0001$;
the dimension of noise vector $z$ is set to $100$; $\lambda_1$, $\lambda_2$ and $\beta$ are set to $0.01$, $5$ and $10$, respectively.
Regarding GCN, the size of the classifier vector is $2,048$; $2$ convolutional layers with a hidden dimension of $2,048$ are used; the learning rate is set to $0.001$.
As for the optimum similarity threshold for creating semantic graphs, we provide a detailed evaluation in Section~\ref{ablation_studies}.

For baselines DisenE and DisenKGAT, we test different $K$ values and report the better ones in the main body, and attach the complete results in Appendix~\ref{disenkge_sensitivity_study}.
More details please see our released codes. 


\noindent\textbf{Overall Results.}
The results are shown in the left side of Table~\ref{tab:overall_results}.
We can see DOZSL always achieves the best performance on AwA and ImNet-O, no matter what downstream ZSL learners  are applied (+GAN or +GCN).
On ImNet-A, DOZSL is still the best in most cases. 
Although DOZSL does not outperform RGCN-ZSL on the metric of $H$, the result is still comparable.

\noindent \textbf{Results on Ontology Encoders.}
First, we find the methods with our disentangled embeddings often outperform those methods with non-disentangled embeddings.
In particular, DOZSL(AGG) outperforms RGAT and TransE on all the datasets no matter what ZSL learners are used.
Second, we find DOZSL(AGG) often performs better than DOZSL(RD) on most metrics.
This indicates the superiority of capturing neighborhood information in learning disentangled ontology embeddings.
Third, our property guided component-wise triple score is quite effective in learning disentangled embeddings.
This can be verified by the fact that DOZSL(AGG) outperforms DisenE and DisenKGAT on all the three datasets. 
Even without aggregation, DOZSL(RD) is still quite good in most cases.

\noindent \textbf{Results on ZSL Learners.}
Using either GAN or GCN can make our framework perform better than the baselines.
Especially, when the input ontology embedding is fixed, we can often select one of them for better performance. 
For example, on AwA, \textit{i)} DOZSL(RD+GAN) has worse performance than DisenE+GAN and DisenE+GCN, but DOZSL(RD+GCN) outperforms DisenE+GCN and DisenE+GAN; \textit{ii)} using GCN with DOZSL(AGG) can achieve good performance, while using GAN with DOZSL(AGG) achieves even higher performance on both metrics $H$ and $acc$.
Moreover, our DOZSL variants with GCN perform better than previous propagation-based ZSL methods in most situations, illustrating that our method can more effectively capture the structural class relationships in ontologies.  



\begin{table*}
\small
\caption{Results ($\%$) of ablation studies with GAN.
The better results in each group are in bold.
}
\label{tab:ablation_ontology_encoder_gen}
\vspace{-0.25cm}
\begin{tabular}{l|p{0.6cm}<{\centering}p{0.6cm}<{\centering}|p{0.6cm}<{\centering}p{0.6cm}<{\centering}|p{0.6cm}<{\centering}p{0.6cm}<{\centering}|p{0.6cm}<{\centering}p{0.6cm}<{\centering}p{0.6cm}<{\centering}p{0.6cm}<{\centering}|p{0.6cm}<{\centering}p{0.6cm}<{\centering}p{0.6cm}<{\centering}p{0.6cm}<{\centering}}
\hline
\multicolumn{1}{c|}{\multirow{2}{*}{\bf Methods}}  & \multicolumn{2}{c}{\bf AwA}   & \multicolumn{2}{c}{\bf ImNet-A} & \multicolumn{2}{c|}{\bf ImNet-O} &
\multicolumn{4}{c}{\bf NELL-ZS} & \multicolumn{4}{c}{\bf Wiki-ZS}
\\
& $acc$ & $H$ & $acc$ & $H$ & $acc$ & $H$ & $hit@10$ & $hit@5$ & $hit@1$ & $MRR$ & $hit@10$ & $hit@5$ & $hit@1$ & $MRR$ \\
\hline
DOZSL(RD+GAN)
& 52.35 & 46.91 & \textbf{37.12} & \textbf{30.18}
& \textbf{34.48} & \textbf{28.57} 
& \textbf{36.4} & \textbf{29.9} & \textbf{16.5}  & \textbf{23.4}
& \textbf{27.9} & \textbf{22.7} & 14.0 & \textbf{19.0}
\\
$\text{DOZSL(RD}_{\text{atten}}$+GAN)
& \textbf{59.40} & \textbf{49.47} & 33.60 & 29.96
& 31.62 & 26.69 
& 34.8 & 29.1 & 15.3 & 22.2 
& 27.2 & 22.0 & \textbf{14.2} & 18.8 
\\
\hline
DOZSL(AGG+GAN)
& \textbf{66.36} & \textbf{57.62}
& \textbf{40.26} & \textbf{32.82}
& \textbf{36.00} & 28.74 
& \textbf{36.2} & 29.5 & \textbf{16.1} & \textbf{23.0}
& \textbf{27.7} & \textbf{22.7} & \textbf{13.3} & \textbf{18.6}
\\
$\text{DOZSL(AGG}_{\text{atten}}$+GAN)
& 61.51 & 51.06
& 34.34 & 30.67
& 30.71 & 26.72
& 35.8 & \textbf{29.6} & 15.9 & 22.9
& 26.7 & 21.8 & 13.0 & 18.1
\\
$\text{DOZSL(AGG}_{\text{sub}}$+GAN)
& 61.29 & 50.65
& 34.93 & 28.45 
& 35.46 & \textbf{29.40}
& 35.7 & 29.0 & 15.2 & 22.3
& 27.1 & 21.8 & 12.6 & 17.9 
\\
\hline
\end{tabular}
\end{table*}

\subsubsection{ZS-KGC}
For ontology encoder, we re-use the settings in ZS-IMGC. The dimension of component embedding and property embedding is set to $200$.
$K$ is $4$ for DOZSL(RD) and is $9$ for DOZSL(AGG) considering the reverse properties and the self-connection property.
The feature extractors are pre-trained to extract $200$-dimensional and $100$-dimensional relation features for NELL-ZS and Wiki-ZS, respectively, following the settings in \cite{qin2020zsgan,geng2021ontozsl}, with TransE-based embeddings as the input.
%
For ZSL learner, we also employ the same GAN and GCN architectures as in ZS-IMGC, but use some different settings.
Regarding the GAN for NELL-ZS, the generator has $250$ hidden units,
while the discriminator has $200$ hidden units.
Regarding the GAN for Wiki-ZS, the corresponding unit numbers are $200$ and $100$.
For both datasets, the noise vector size is set to $15$; $\lambda_1$, $\lambda_2$ are set to $1$ and $3$, respectively.
Regarding GCN, the classifier vector size is $200$ for NELL-ZS and $100$ for Wiki-ZS.
As in ZS-IMGC, the selection of similarity thresholds for creating semantic graphs is evaluated in Section~\ref{ablation_studies}; different $K$ values are tested for DisenE and DisenKGAT with the optimum performance reported in Table \ref{tab:overall_results} and the complete results attached in Appendix \ref{disenkge_sensitivity_study}.

\noindent \textbf{Overall Results.}
The results are presented in the right of Table~\ref{tab:overall_results}.
On NELL-ZS, our method achieves the best on $hit@10$ and $hit@5$, DOZSL(RD+GAN) and DOZSL(RD+GCN) are both very competitive to the baseline RGCN-ZSL and better than other baselines on $hit@1$ and $MRR$.
On Wiki-ZS, two baselines RGAT+GAN and RGCN-ZSL perform the best, but our method DOZSL(RD+GAN) is very close to them, especially on $MRR$ ($19.1$ vs $19.0$) and $hit@1$ ($14.2$ vs $14.0$).


\noindent \textbf{Results on Ontology Encoders.}
It can be observed that the performance gap between DOZSL(RD) and DOZSL(AGG) is narrowed, and DOZSL(RD) even performs better on some metrics, which can be attributed to the following reasons.
(1) The neighborhood information of concepts in ZS-IMGC task is richer than that in ZS-KGC task, especially for the concepts in NELL-ZS's ontology.
Statistically, the average number of surrounding neighbors for NELL-ZS is around $3.4$, while the number for ImNet-A is around $5.9$.
(2) The properties in the ontologies of ZS-IMGC task such as \textit{imgc:hasAttribute} are 1-N; while most properties in the ontologies of ZS-KGC task are 1-1.
The embedding methods that ignore aggregating the neighborhood are often not good at handling these 1-N properties.
Besides, in comparison with the disentangled and non-disentangled baselines, our methods always have superior performances, i.e., on most metrics, DOZSL(AGG) performs better than RGAT and DisenKGAT, and DOZSL(RD) outperforms DisenE and TransE by a large margin.

\noindent \textbf{Results on ZSL Learners.}
Given the same ontology embeddings, we find the performance varies from one ZSL learner to another.
The GCN-based learner usually performs better than the GAN-based one on NELL-ZS, while the GAN-based learner reversely performs better on Wiki-ZS.
This motivates us to conduct an in-depth analysis about the interaction between the datasets and the ZSL methods, so that making a more suitable selection for better performance.
Moreover, RGCN-ZSL also shows the promising ability of relation-aware GNNs on ZS-KGC task.


\vspace{-0.2cm}
\subsection{Ablation Studies}\label{ablation_studies}

We conduct extensive ablation studies to analyze the impact of different factors in DOZSL, including the property guided triple scoring, the neighborhood aggregation, the similarity threshold for constructing semantic graphs and the classifier fusion.

\noindent\textbf{Property Guided Triple Scoring.}
We replace the property guided triple scoring in DOZSL(RD) and DOZSL(AGG) by the widely-adopted attentive triple scoring and keep the same setting of $K$.
This leads to two new variants, denoted as $\text{DOZSL(RD}_{\text{atten}})$ and $\text{DOZSL(AGG}_{\text{atten}})$, respectively.
These variants' results with GAN are reported in Table~\ref{tab:ablation_ontology_encoder_gen}, the results with GCN are attached in Appendix~\ref{ablation_ontology_encoder}.
We can find that $\text{DOZSL(RD}_{\text{atten}})$ and $\text{DOZSL(AGG}_{\text{atten}})$ always obtain dramatically worse results than $\text{DOZSL(RD})$ and $\text{DOZSL(AGG})$, respectively, on all the datasets of the two tasks, with the only exception of $\text{DOZSL(RD}_{\text{atten}}$+GAN) on AwA.
These results illustrate the effectiveness of our proposed property guided triple scoring.
The except may be due to the imbalanced associated triples of different properties in AwA's ontology: \textit{imgc:hasAttribute} has $1,562$ associated triples, which can well train its corresponding component, while \textit{rdfs:subClassOf} has only $197$ associated triples, making its corresponding component under fitted.
The two components are concatenated and fed to GANs together, thus they may influence each other.
In contrast, the GCN-based method, which performs independent feature propagation in isolated semantic graphs, suffers less from the imbalance issue.

\noindent\textbf{Neighborhood Aggregation.}
In DOZSL, we aggregate information from all the neighboring concepts in the ontology, with an attention mechanism for combination.
Here, we want to test a more straightforward solution, i.e., aggregating information from a neighborhood subset which only includes concepts that are connected by the property corresponding to the embedding component.
This leads to new variants denoted by $\text{DOZSL(AGG}_{\text{sub}})$.
The results with GAN are shown in Table~\ref{tab:ablation_ontology_encoder_gen}, the results with GCN are in Appendix~\ref{ablation_ontology_encoder}.
In comparison with DOZSL (AGG), $\text{DOZSL(AGG}_{\text{sub}})$ performs worse on most metrics across two tasks, except for  $\text{DOZSL}$ $\text{(AGG}_{\text{sub}}$+GAN) on ImNet-O w.r.t. $H$ and $\text{DOZSL(AGG}_{\text{sub}}$+GCN) on NELL-ZS.
The overall worse results of $\text{DOZSL(AGG}_{\text{sub}})$ indicate that learning a component embedding should (attentively) aggregate all the neighboring concepts rather than select a part of them according to the specific properties.
The exceptions may be due to the simple neighborhoods in 
NELL-ZS and ImNet-O and/or the independent propagation in each semantic graph.

\noindent\textbf{Similarity Threshold and Classifier Fusion.}
We compare different similarity thresholds ranging from $0.85$ to $0.999$ for constructing semantic graphs, and compare different classifier fusion functions, under different ontology encoding methods.
The results are reported in Figure~\ref{ablation_study_sim_fusion} in Appendix~\ref{ablation_sim_fusion}, from which we can find that the optimum similarity threshold varies when different ontology encoding methods are used, and the two fusion functions --- Average and Linear Transformation both positively contribute to the learning of the classifier.
Please see Appendix~\ref{ablation_sim_fusion} for more details.

\vspace{-0.2cm}
\subsection{Case Study}
\begin{figure}
\centering  
\includegraphics[width=0.85\linewidth]{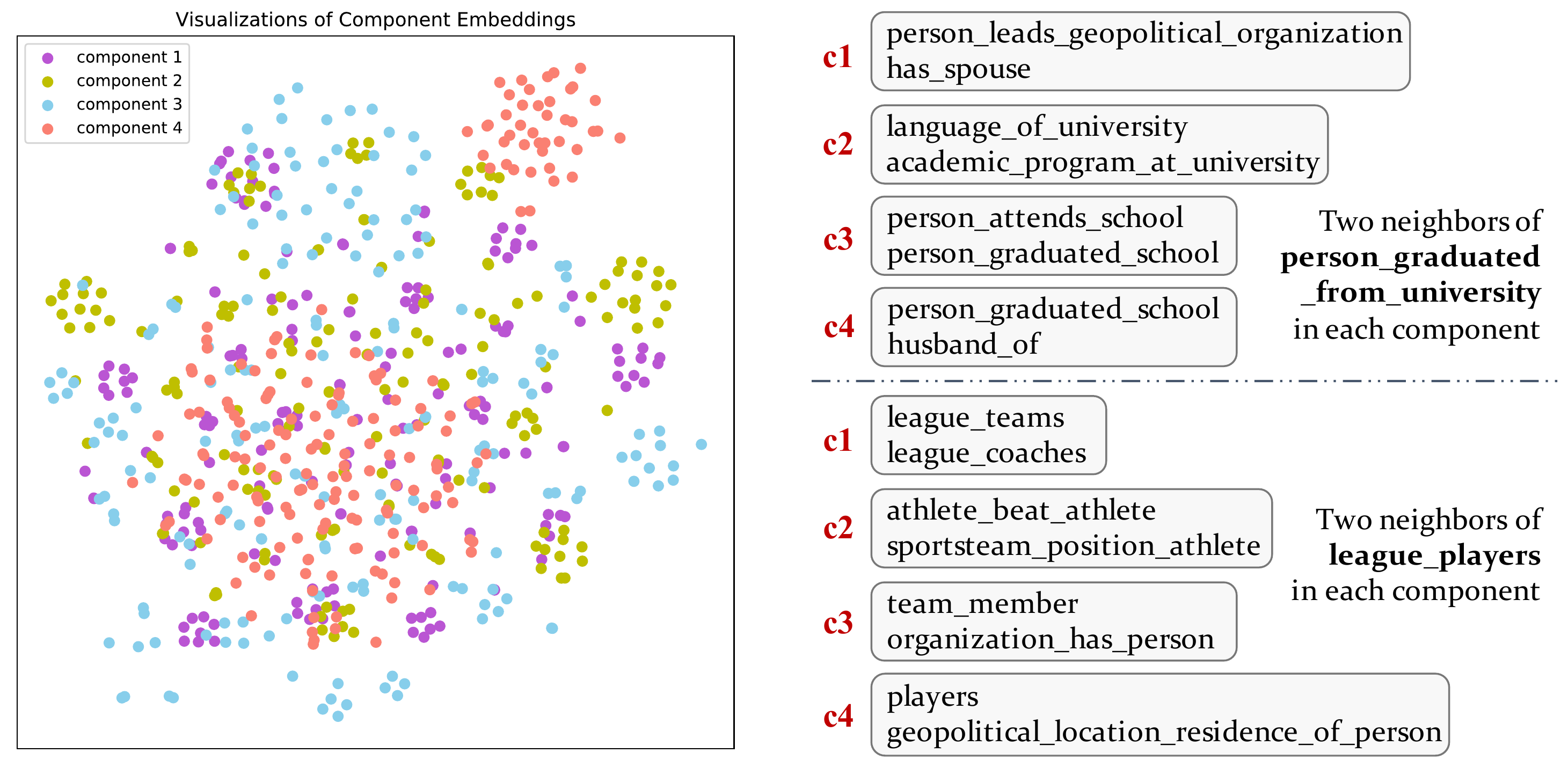}
\vspace{-0.2cm}
\caption{Cases of relations in NELL-ZS. Best viewed in color.}
\label{case_study}

\end{figure}

We use examples from NELL-ZS to analyze disentanglement of concept embeddings we learned.
In the left of Figure~\ref{case_study}, we visualize the component embeddings of KG relations learned from NELL-ZS's ontology by DOZSL(RD), where different colors indicate different components.
We can find that \textit{i)} the embeddings are clustered into different groups under each component's subspace, and \textit{ii)} the component embeddings of each relation are divided into different clusters across different components.
These observations illustrate that \textit{i)} our method indeed captures the semantically similarity among relation concepts under each semantic aspect and \textit{ii)} different relatedness is presented across different aspects.

Also, to further verify that different components represent different semantic aspects, for each relation, we randomly select two neighbors from the cluster of each component.
The right of Figure~\ref{case_study} presents two examples.
For relation \textit{league\_players}, its two neighbors from the first component are \textit{league\_teams} and \textit{league\_coaches}, the head entity types of these three relations are identical, i.e., \textit{sports\_league}; while its two neighbors from the second component are \textit{athlete\_beat\_athlete} and \textit{sports\_team\_position\_athlete}, their tail entity types are \textit{athlete}.
According to these two examples, we can find that these four components respectively reflect four semantic aspects of the relations, i.e., \textit{rdfs:domain}, \textit{rdfs:range}, \textit{rdfs:subPropertyOf} and \textit{rdfs:subClassOf}, and we can also conclude that the semantic of one component is a fixed across different relations. 


\section{Conclusion and Discussion}
In this study, we focused on ontology augmented ZSL and proposed a novel property guided disentangled ontology embedding method.
With the new disentangled embeddings, different semantic aspects of ZSL classes are figured out and more fine-grained inter-class relationships are extracted, through which the ontology can be better utilized.
To integrate these disentangled embeddings, we also developed a general ZSL framework DOZSL, including a GAN-based generative model and a GCN-based propagation model.
Extensive evaluations with ablation studies and case studies on five datasets of ZS-IMGC and ZS-KGC show that DOZSL often outperforms the state-of-the-art baselines and its components are quite effective.

DOZSL is compatible to both ZSL learners developed by us, and they together lead to higher robustness and better performance.
Meanwhile, the performance of DOZSL is less competitive to the state-of-the-art on one of the five datasets.
This motivates us to take an in-depth analysis of this dataset and its ontology, and to develop more robust disentangled embedding methods and ZSL learners in the future.
We also realize some relation-aware GNNs such as RGCN achieve quite promising results on some datasets. This motivates us to study the propagation-based ZSL learner with these GNNs.
Lastly, we will apply and evaluate DOZSL in other tasks such as open information extraction and visual question answering.


\begin{acks}
This work is partially funded by NSFCU19B2027/91846204, 
the EPSRC project ConCur (EP/V050869/1) and 
the Chang Jiang Scholars Program (J2019032).
\end{acks}



\balance
\bibliographystyle{ACM-Reference-Format}
\bibliography{sample-base}


\appendix
\begin{table*}[]
\small
\caption{Results ($\%$) of DisenE and DisenKGAT with GAN w.r.t different $K$ values.
The better results are in bold.
}
\label{tab:disenkge_K_results}
\vspace{-0.3cm}
\begin{tabular}{l|c|p{0.6cm}<{\centering}p{0.6cm}<{\centering}|p{0.6cm}<{\centering}p{0.6cm}<{\centering}|p{0.6cm}<{\centering}p{0.6cm}<{\centering}|p{0.6cm}<{\centering}p{0.6cm}<{\centering}p{0.6cm}<{\centering}p{0.6cm}<{\centering}|p{0.6cm}<{\centering}p{0.6cm}<{\centering}p{0.6cm}<{\centering}p{0.6cm}<{\centering}}
\hline
\multicolumn{1}{c|}{\multirow{2}{*}{\bf Methods}}  & \multicolumn{1}{c|}{\multirow{2}{*}{\bf K}} & \multicolumn{2}{c}{\bf AwA}   & \multicolumn{2}{c}{\bf ImNet-A} & \multicolumn{2}{c|}{\bf ImNet-O} &
\multicolumn{4}{c}{\bf NELL-ZS} & \multicolumn{4}{c}{\bf Wiki-ZS}
\\
& & $acc$ & $H$ & $acc$ & $H$ & $acc$ & $H$ & $hit@10$ & $hit@5$ & $hit@1$ & $MRR$ & $hit@10$ & $hit@5$ & $hit@1$ & $MRR$ \\
\hline
DisenE+GAN & 2
& \textbf{59.40} & \textbf{49.47} & \textbf{33.60} & \textbf{29.96}
& \textbf{31.62} & \textbf{26.69} 
& 34.5 & 28.2 & \textbf{15.4}  & 22.0
& \textbf{28.0} & \textbf{22.7} & 13.8 & \textbf{18.9} 
\\
DisenE+GAN & 4
& 44.59 & 41.69 & 24.27 & 23.79 
& 21.62 & 21.80 
& \textbf{34.8} & \textbf{29.1} & 15.3 & \textbf{22.2} 
& 27.2 & 22.0 & \textbf{14.2} & 18.8 
\\
\hline
DisenKGAT+GAN & 2
& 60.20 & 54.08
& \textbf{35.90} &  \textbf{31.09} 
& \textbf{34.94} & \textbf{27.33}  
& \textbf{35.9} & \textbf{29.5} & 15.7 & \textbf{22.9}
& 27.5 & 22.0 & 13.7 & 18.6 
\\
DisenKGAT+GAN & 4
& \textbf{61.81}& \textbf{54.41}
& 31.35 & 28.57 
& 31.58 & 27.13 
& 35.0 & 28.7 & \textbf{16.1} & 22.5
& \textbf{27.6} & \textbf{22.4} & \textbf{13.8} & \textbf{18.8}
\\
\hline
\end{tabular}
\end{table*}

\begin{table*}
\small
\caption{Results ($\%$) of ablation studies with GCN. The better results in each group are in bold.
}
\label{tab:ablation_ontology_encoder_pro}
\vspace{-0.3cm}
\begin{tabular}{l|p{0.6cm}<{\centering}p{0.6cm}<{\centering}|p{0.6cm}<{\centering}p{0.6cm}<{\centering}|p{0.6cm}<{\centering}p{0.6cm}<{\centering}|p{0.6cm}<{\centering}p{0.6cm}<{\centering}p{0.6cm}<{\centering}p{0.6cm}<{\centering}|p{0.6cm}<{\centering}p{0.6cm}<{\centering}p{0.6cm}<{\centering}p{0.6cm}<{\centering}}
\hline
\multicolumn{1}{c|}{\multirow{2}{*}{\bf Methods}}  & \multicolumn{2}{c}{\bf AwA}   & \multicolumn{2}{c}{\bf ImNet-A} & \multicolumn{2}{c|}{\bf ImNet-O} &
\multicolumn{4}{c}{\bf NELL-ZS} & \multicolumn{4}{c}{\bf Wiki-ZS}
\\
& $acc$ & $H$ & $acc$ & $H$ & $acc$ & $H$ & $hit@10$ & $hit@5$ & $hit@1$ & $MRR$ & $hit@10$ & $hit@5$ & $hit@1$ & $MRR$ \\
\hline
DOZSL(RD+GCN)
& \textbf{62.79} & \textbf{52.74}
& \textbf{36.01} & \textbf{30.29} 
& \textbf{33.66} & \textbf{31.19} 
& \textbf{38.0} & \textbf{31.2} & \textbf{16.5} & \textbf{23.9} 
& \textbf{26.7} & \textbf{21.9} & \textbf{13.8} & \textbf{18.5} 
\\
$\text{DOZSL(RD}_{\text{atten}}$+GCN)
& 58.34 & 50.86 
& 32.56 & 27.76 
& 32.02 & 26.33
& 35.5 & 29.7 & 15.6 & 22.7  
& 26.3 & 21.4 & 13.4 & 18.1 
\\
\hline
DOZSL(AGG)+PRO
& \textbf{63.88} & \textbf{44.52} 
& \textbf{38.69} & \textbf{32.12} 
& \textbf{37.42} & \textbf{31.77} 
& 36.2 & 29.3 & \textbf{16.2} & 23.0 
& \textbf{27.5} & \textbf{22.4} & 13.6 & \textbf{18.7} 
\\
$\text{DOZSL(AGG}_{\text{atten}}$+GCN)
& 54.40 & 32.00 
& 36.18 & 27.55 
& 31.47 & 26.54 
& 35.3 & 28.8 & 15.6 & 22.3 
& 27.5 & 22.1 & \textbf{13.7} & 18.7   
\\
$\text{DOZSL(AGG}_{\text{sub}}$+GCN)
& 63.66 & 33.19 
& 35.03 & 26.63 
& 35.37 & 31.11 
& \textbf{36.9} & \textbf{30.0} & 15.8 & \textbf{23.1} 
& 27.1 & 21.8 & 13.4 & 18.3 \\
\hline
\end{tabular}
\end{table*}

\begin{figure*}
\centering  
\vspace{-0.35cm} 
\subfigbottomskip=-2pt 
\subfigure[Accuracy on AwA]{
\label{level.sub.1}
\includegraphics[width=0.25\linewidth]{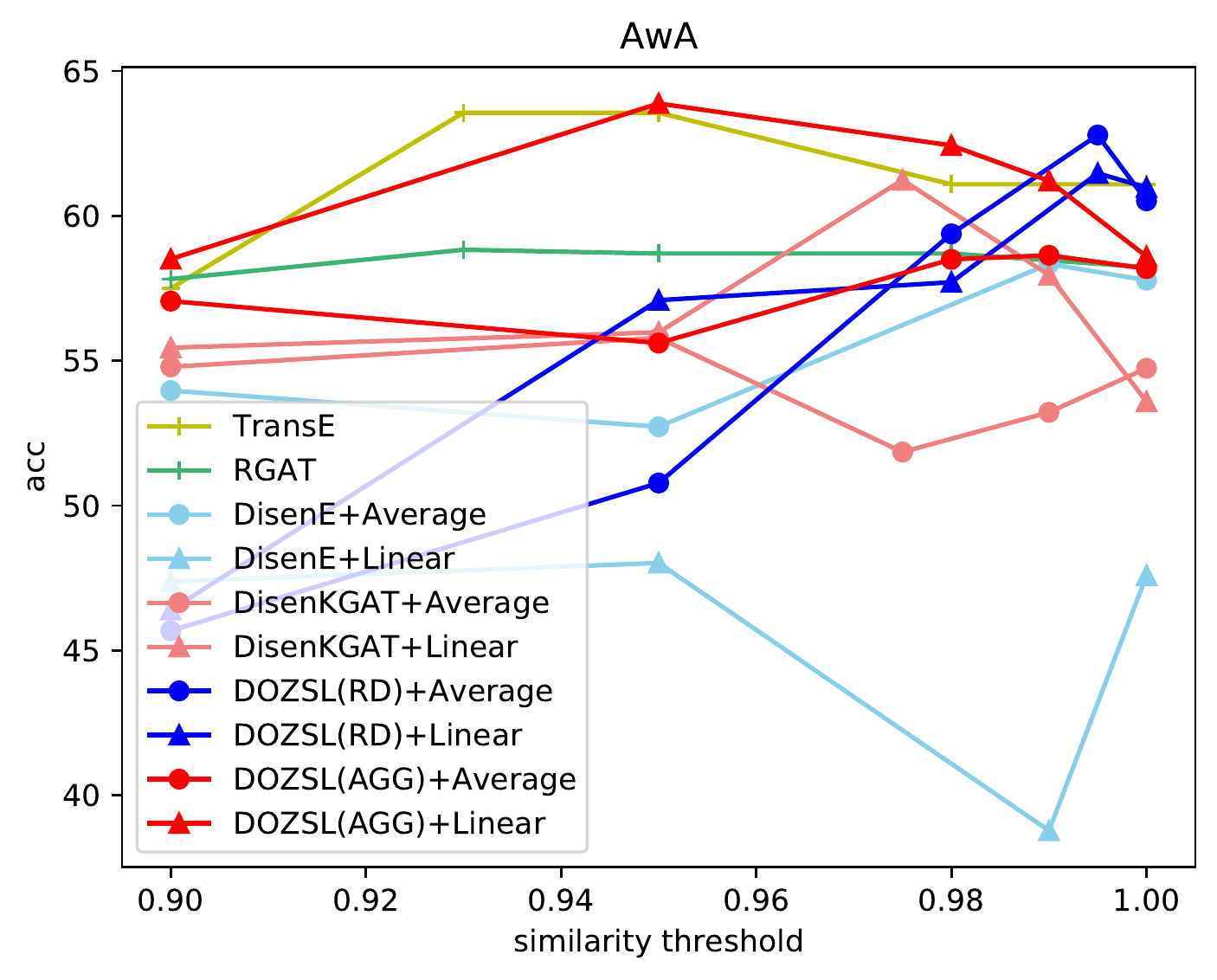}}
\quad 
\subfigure[Accuracy on ImNet-A]{
\label{level.sub.2}
\includegraphics[width=0.25\linewidth]{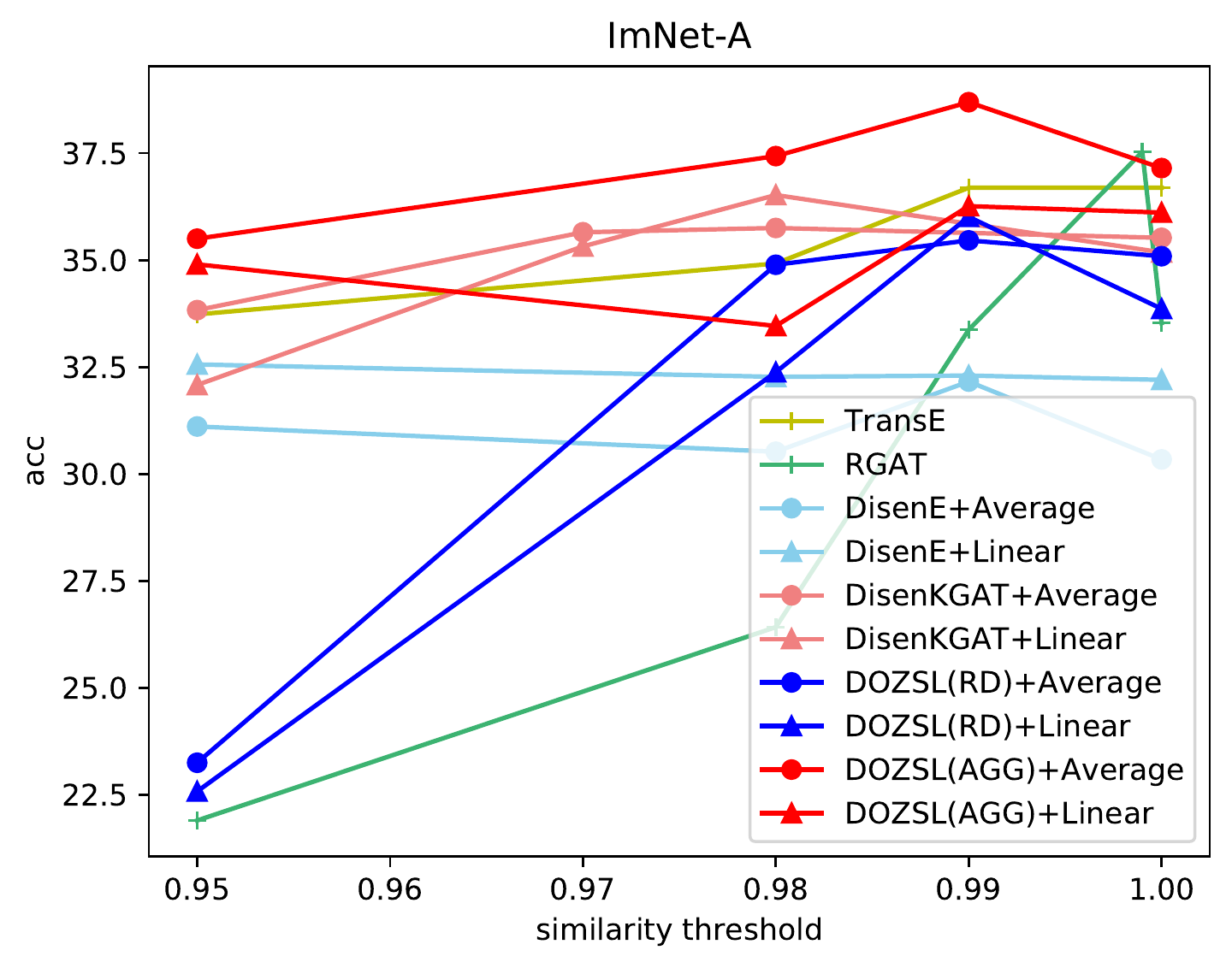}}
\quad
\subfigure[Accuracy on ImNet-O]{
\label{level.sub.3}
\includegraphics[width=0.25\linewidth]{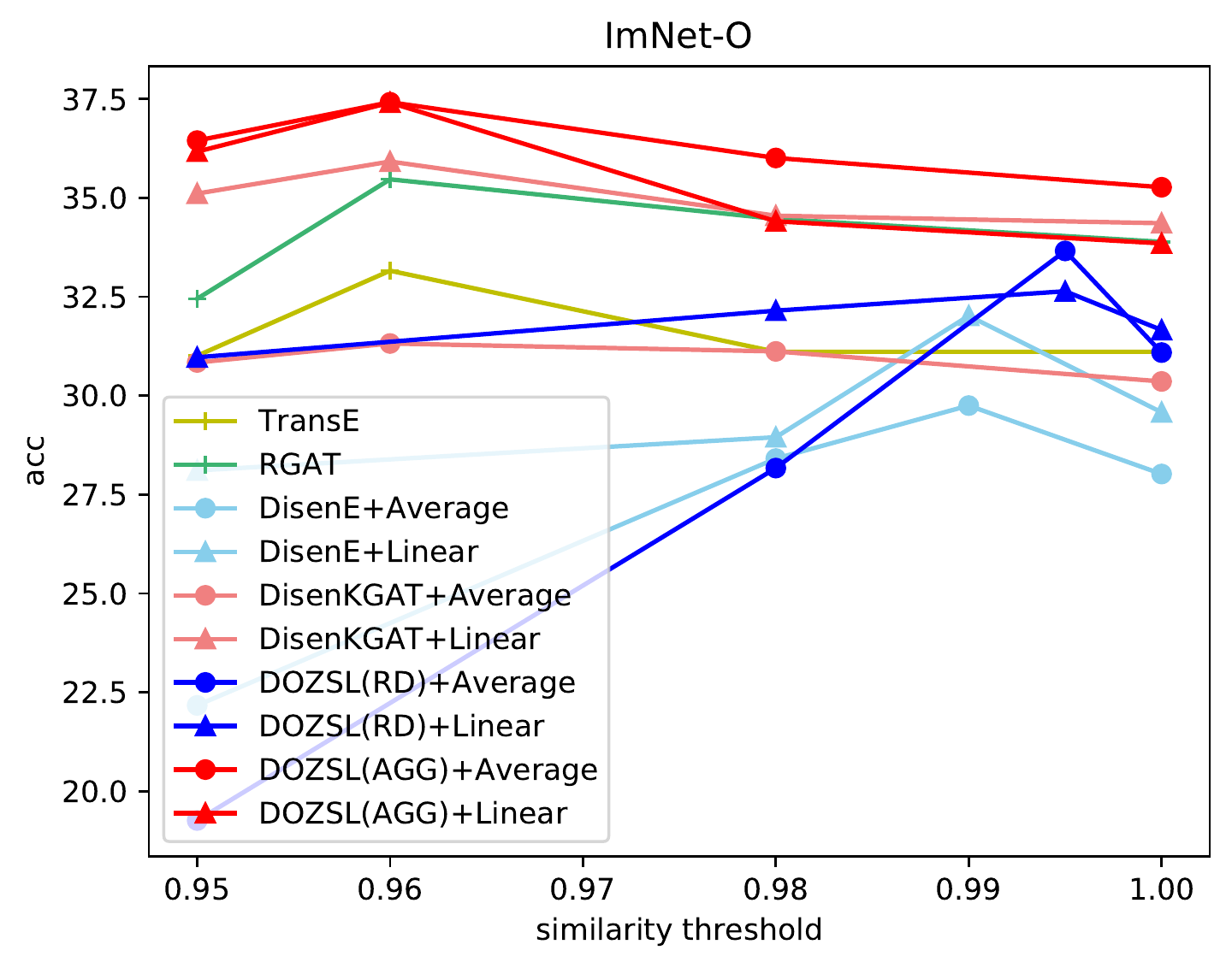}}
\\
\subfigure[Hit@10 and MRR on NELL-ZS]{
\label{level.sub.4}
\includegraphics[width=0.26\linewidth]{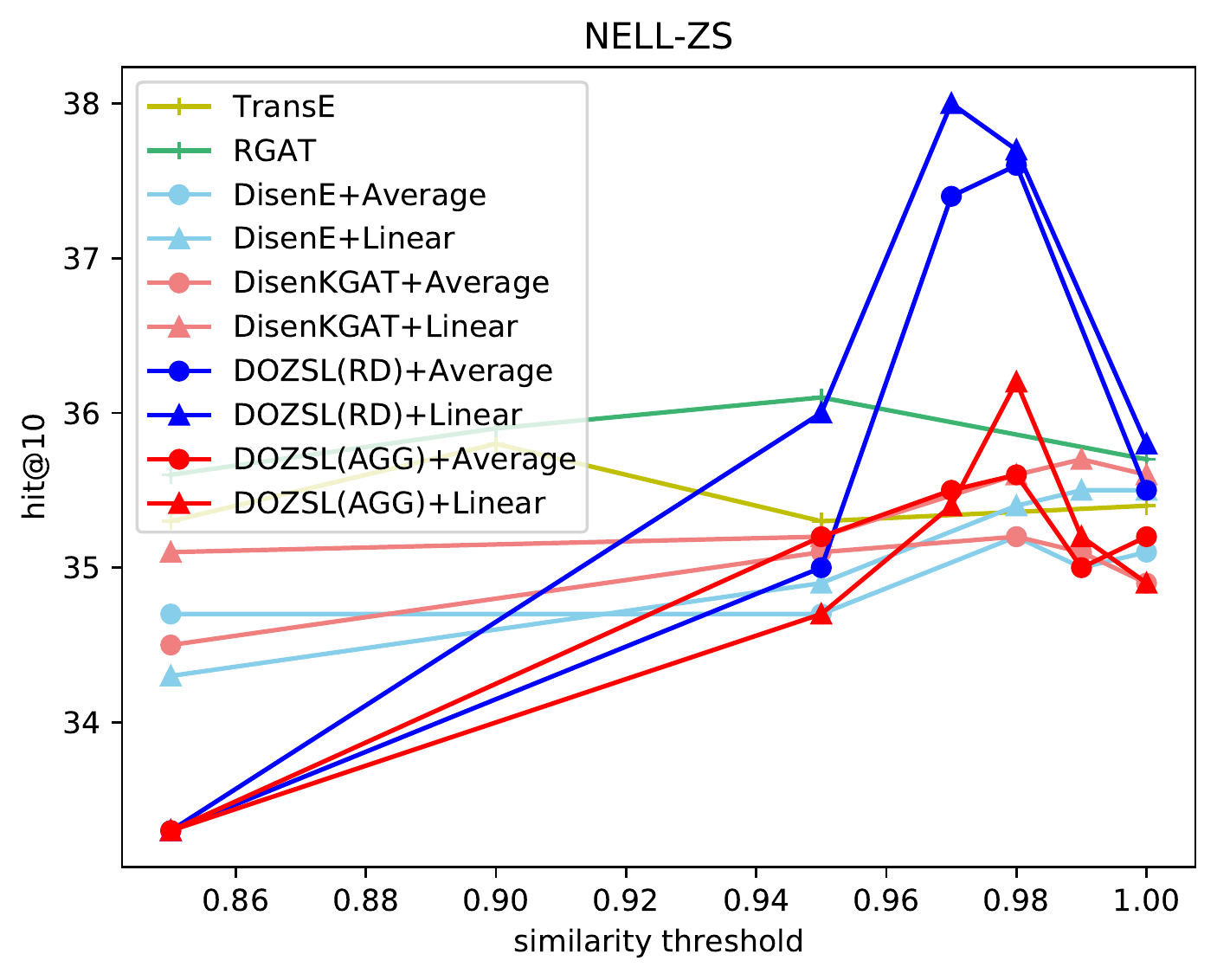}
\includegraphics[width=0.26\linewidth]{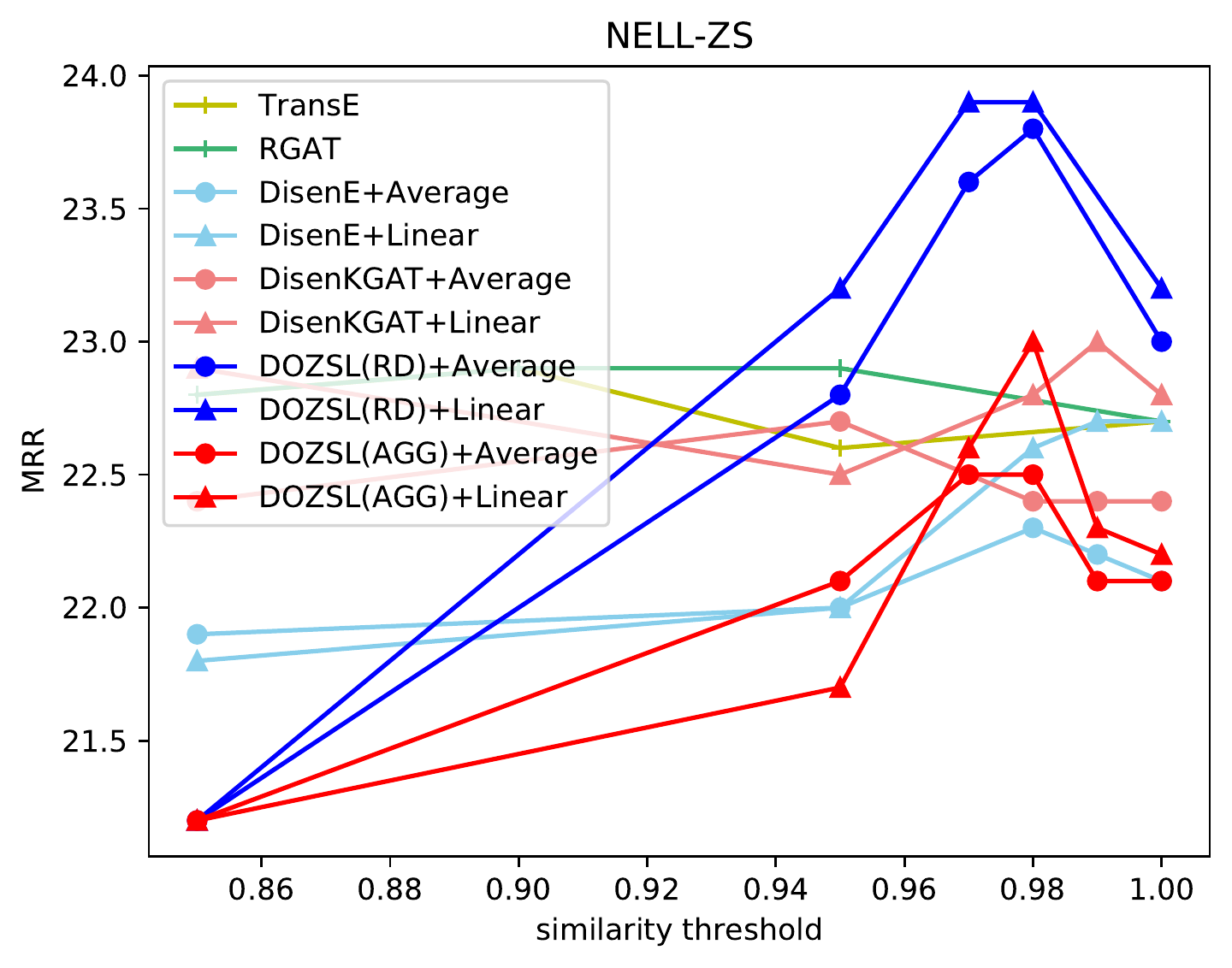}
}
\quad
\subfigure[Hit@10 and MRR on Wiki-ZS]{
\label{level.sub.6}
\includegraphics[width=0.26\linewidth]{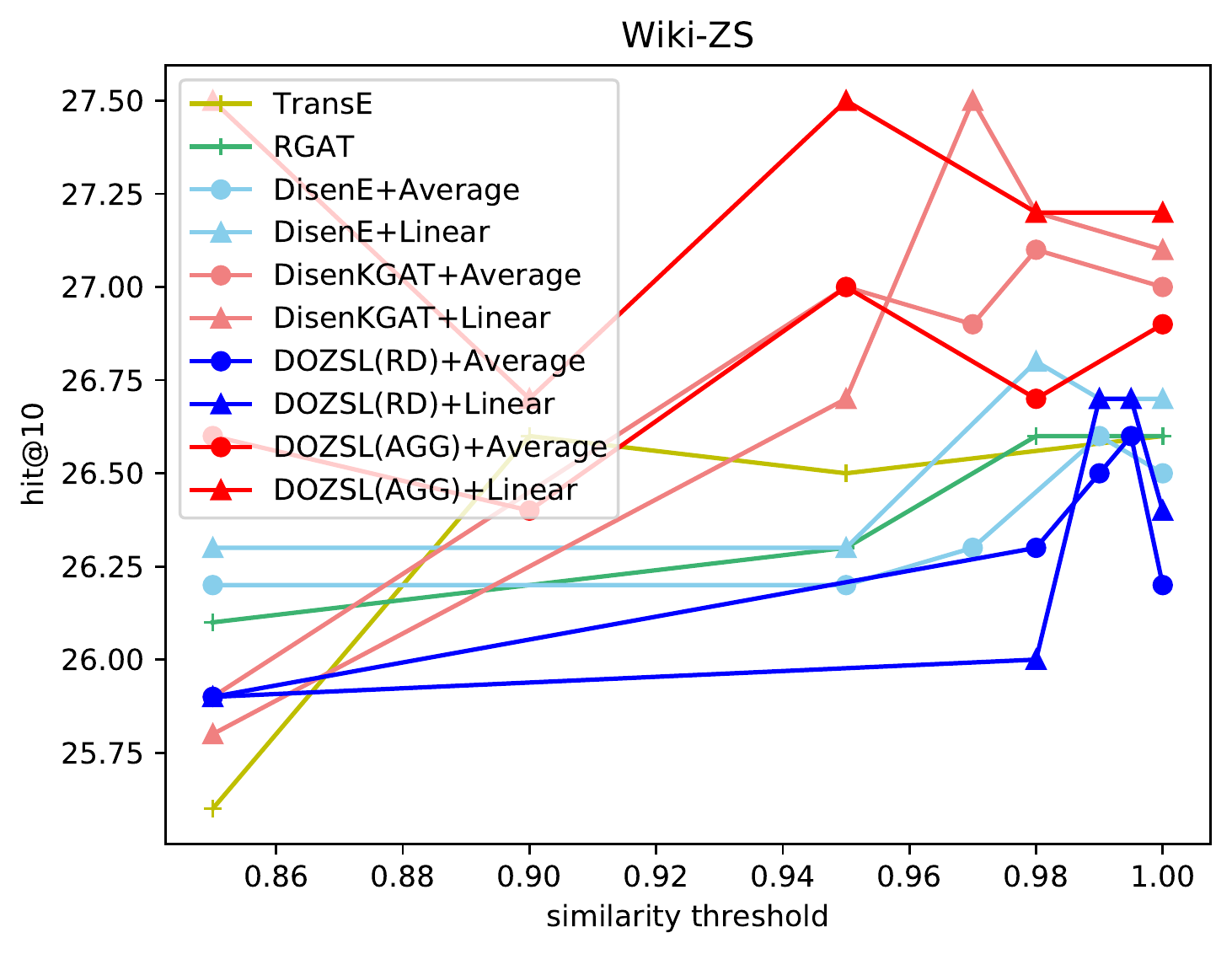}
\includegraphics[width=0.26\linewidth]{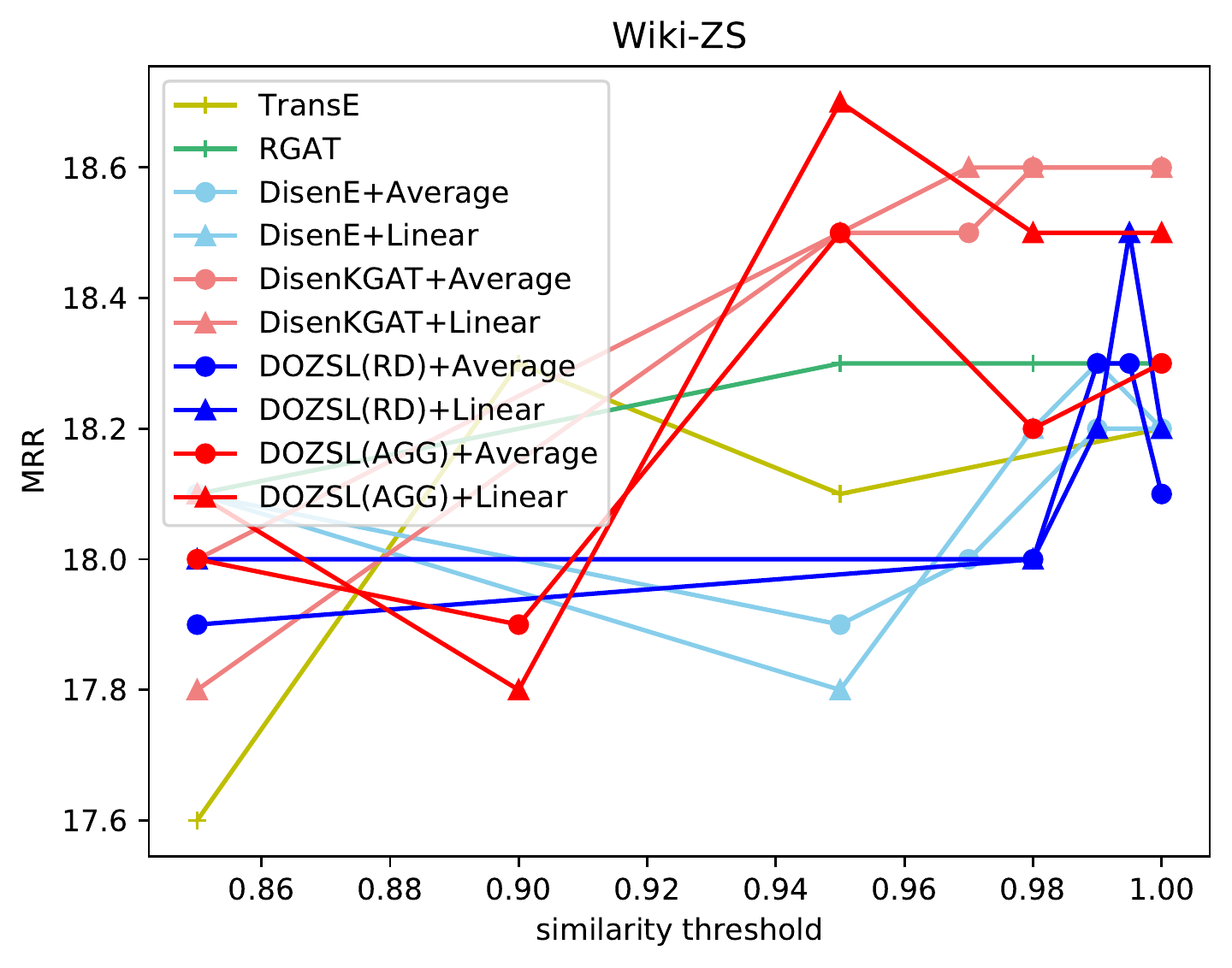}
}
\vspace{-0.2cm}
\caption{Results of GCN-based DOZSL variants using different ontology encoders with different similarity thresholds and different classifier fusion functions. Best viewed in color.}
\label{ablation_study_sim_fusion}
\end{figure*}

\section{Sensitivity Study of DisenE and DisenKGAT}\label{disenkge_sensitivity_study}
In this section, we study the sensitivity of the number of components $K$ used in the baselines DisenE \cite{kou2020disene} and DisenKGAT \cite{wu2021disenkgat}.
Specifically, we $K$ to $2$ and $4$, two values with which the baselines perform  well, and experiment with the GAN-based learner.
The results on the six datasets of the two ZSL tasks are presented in Table~\ref{tab:disenkge_K_results}.
We can find that DisenE gets higher performance on all the three ZS-IMGC datasets and on Wiki-ZS when $K=2$.
It also gets better results on most metrics on NELL-ZS when $K=4$.
As for DisenKGAT, the optimum $K$ values on AwA, ImNet-A, ImNet-O, NELL-ZS and Wiki-ZS are $4, 2, 2, 2, 4$, respectively.

\section{Ablation Study of The Ontology Encoder with GCN-based methods}\label{ablation_ontology_encoder}
In this section, we report the results of ablation studies on the property guided triple scoring and the neighborhood aggregation in the disentangled ontology encoder when incorporating with GCN-based methods.
The results are shown in Table~\ref{tab:ablation_ontology_encoder_pro}.

\section{Ablation Study of The GCN-based Learner}\label{ablation_sim_fusion}
In this section, we study the impact of the similarity threshold and the classifier fusion function under different disentangled ontology embeddings, using all our evaluation datasets.
The results are presented in Figure~\ref{ablation_study_sim_fusion}.
Specifically, we report the results of the metric of $acc$ (i.e., the standard ZSL testing setting) for ZS-IMGC task and the results of the metrics of $hit@10$ and $MRR$ for ZS-KGC task.
Moreover, the curve of the Average fusion function is decorated with circular, while the curve of the Linear Transformation fusion function is decorated with triangle.
Different ontology encoding methods are presented in different colors.

\end{document}